\documentclass[sigconf,natbib=true,anonymous=false,table]{acmart}
\usepackage{multirow} 
\usepackage{graphicx}    
\usepackage{subcaption}
\usepackage{algorithm}
\usepackage{algorithmic}

\usepackage{enumitem}

\AtBeginDocument{%
  }

\copyrightyear{2025}
\acmYear{2025}
\setcopyright{acmlicensed}\acmConference[SIGIR '25]{Proceedings of the 48th International ACM SIGIR Conference on Research and Development in Information Retrieval}{July 13--18, 2025}{Padua, Italy}
\acmBooktitle{Proceedings of the 48th International ACM SIGIR Conference on Research and Development in Information Retrieval (SIGIR '25), July 13--18, 2025, Padua, Italy}
\acmDOI{10.1145/3726302.3729930}
\acmISBN{979-8-4007-1592-1/2025/07}

\renewcommand{\shortauthors}{}

\begin{document}

\title{CoMaPOI: A Collaborative Multi-Agent Framework for Next POI Prediction Bridging the Gap Between Trajectory and Language}






\author{Lin Zhong}
\email{zhonglin@stu.hit.edu.cn}
\affiliation{
  \institution{Department of Computer Science and Technology of Harbin Institute of Technology, Shenzhen}
  \city{Shenzhen}
  \state{Guangdong}
  \country{China}
}

\author{Lingzhi	Wang}
\email{wanglingzhi@hit.edu.cn}
\affiliation{
  \institution{Department of Computer Science and Technology of Harbin Institute of Technology, Shenzhen}
  \city{Shenzhen}
  \state{Guangdong}
  \country{China}
}

\author{Xu Yang}
\email{xuyang97@stu.hit.edu.cn}
\affiliation{
  \institution{Department of Computer Science and Technology of Harbin Institute of Technology, Shenzhen}
  \city{Shenzhen}
  \state{Guangdong}
  \country{China}
}

\author{Qing Liao}
\email{liaoqing@hit.edu.cn}
\affiliation{
  \institution{Department of Computer Science and Technology of Harbin Institute of Technology, Shenzhen}
  \city{Shenzhen}
  \state{Guangdong}
  \country{China}
}

\renewcommand{\shortauthors}{Lin Zhong, Lingzhi	Wang, Xu Yang, and Qing Liao}

\authornote{Corresponding author: Qing Liao, Email: liaoqing@hit.edu.cn}



\begin{abstract}

Large Language Models (LLMs) offer new opportunities for the next Point-Of-Interest (POI) prediction task, leveraging their capabilities in semantic understanding of POI trajectories. However, previous LLM-based methods, which are superficially adapted to next POI prediction, largely overlook critical challenges associated with applying LLMs to this task. Specifically, LLMs encounter two critical challenges: (1) a lack of intrinsic understanding of numeric spatiotemporal data, which hinders accurate modeling of users’ spatiotemporal distributions and preferences; and (2) an excessively large and unconstrained candidate POI space, which often results in random or irrelevant predictions. To address these issues, we propose a \underline{Co}llaborative \underline{M}ulti-\underline{A}gent Framework for Next POI Prediction, named \textit{CoMaPOI}. Through the close interaction of three specialized agents (Profiler, Forecaster, and Predictor), CoMaPOI collaboratively addresses the two critical challenges. The Profiler agent is responsible for converting numeric data into language descriptions, enhancing semantic understanding. The Forecaster agent focuses on dynamically constraining and refining the candidate POI space. The Predictor agent integrates this information to generate high-precision predictions. Extensive experiments on three benchmark datasets (NYC, TKY, and CA) demonstrate that CoMaPOI achieves state-of-the-art performance, improving all metrics by 5\% to 10\% compared to SOTA baselines. This work pioneers the investigation of challenges associated with applying LLMs to complex spatiotemporal tasks by leveraging tailored collaborative agents\footnote{Our source code is available at: https://github.com/Chips98/CoMaPOI.}.

\end{abstract}






\maketitle

\section{Introduction}

With the rapid development of mobile internet and positioning technologies, next Point-of-Interest (POI) prediction has become a core technique in smart cities, travel planning, and commercial applications \cite{poi1, poi2}. By modeling users' mobility preferences from their check-in trajectories, POI prediction can generate the most likely location recommendations for the user's next movement \cite{zhao2020go, lai2024disentangled, feng2024move}. Traditional deep learning-based POI prediction methods have made notable progress, such as leveraging sequence models and attention mechanisms to capture spatiotemporal features \cite{kang2018self, dang2023uniform, liu2024cbrec}, or incorporating contrastive learning, graph neural networks, and diffusion models to enrich multidimensional data representations \cite{zeng2025global, zhong2024scfl, yin2023next}. However, these methods still face challenges in semantic understanding and interpretability when dealing with complex and dynamic user demands \cite{yang2024siamese, wang2024embracing}. Therefore, how to effectively extract semantic information from trajectory data to further construct more accurate user mobility preferences remains a key challenge in POI prediction tasks.



Recently, Large Language Models (LLMs) have demonstrated powerful generation and contextual reasoning abilities in natural language processing tasks \cite{bao2023large, tan2024idgenrec}. Researchers have begun exploring how to integrate LLMs into spatiotemporal tasks such as POI, traffic management \cite{KOLAT2023100102}, and flow prediction \cite{WANG2023308}. By leveraging LLMs’ strengths in semantic representation and understanding, researchers convert sequential data or POI attributes into natural language descriptions and further extract deeper features, thereby improving the effectiveness of downstream tasks \cite{wei2024llmrec, kim2024large, geng2024breaking}. For example, LLM4POI \cite{li2024large} extracts consensus information among similar users with the help of LLMs, aiding in the next POI prediction for the target user. \citet{kim2024poi} proposes POI GPT, which attempts to leverage LLMs to extract more accurate check-in data from social media. Fundamentally, these methods demonstrate how LLMs can be used to enhance the utility of POI data, showcasing their potential in POI prediction. However, although LLM-based methods have shown potential in data enhancement and semantic representation for POI prediction, they often do not fully consider two core limitations.

Firstly, LLMs inherently lack the ability to comprehend numerical spatiotemporal data, such as POI trajectories \cite{feng2024move}. Since their pre-training is primarily based on textual data, these models struggle with understanding the physical meanings of coordinates, time, and distance. This limitation hinders their capacity to accurately capture user mobility preferences \cite{liu2024can}. For instance, a user's check-ins might show locations with similar latitude and longitude values that are far apart in reality, such as different floors in a skyscraper or locations separated by a river. Moreover, LLMs lack the intrinsic spatiotemporal understanding needed to discern these subtleties. This limitation becomes more pronounced when faced with irregular check-in intervals, like visits to seasonal attractions, or when navigating diverse POI categories, each with unique spatial contexts.

Secondly, when directly using LLMs for POI prediction, the candidate space is vast. For example, there may be tens of thousands of POIs in city, making it impractical and costly to provide all this information to LLMs. Although some researchers have attempted to modify the output structures of LLMs using embedding vectors, these approaches may weaken linguistic capabilities and increase the complexity of applications \cite{jin2023time, yuan2024back}. Therefore, without altering the structure of LLMs, a more feasible approach is to provide a refined candidate set that includes high-probability options. This set should maximize the inclusion of the true label while minimizing its size to reduce noise from irrelevant POIs. Therefore, providing a high-quality candidate POI set to reduce prediction errors is the second challenge this paper aims to address.

To address the two challenges outlined above, this paper proposes a \textbf{\underline{Co}llaborative \underline{M}ulti-\underline{A}gent Framework for Next POI Prediction (CoMaPOI)}. Inspired by research on multi-agent collaboration \cite{chan2023chateval, wang2024rethinking}, CoMaPOI addresses the complex POI prediction task by decomposing it into multiple subtasks. CoMaPOI leverages the complementary strengths of multiple agents, each built on an LLM, to handle specific aspects of the task. Specifically, CoMaPOI incorporates three agents. The \textbf{Profiler} agent transforms users’ trajectories into natural language descriptions, allowing the LLM to better understand users’ profiles and mobility patterns. The \textbf{Forecaster} agent is responsible for generating candidate sets optimized based on mobility preferences, thus providing a more precise and narrowed set of options for prediction. Finally, the \textbf{Predictor} agent integrates user profiles, mobility patterns, and the refined candidate sets to produce more accurate next POI predictions. During the fine-tuning phase, to better adapt CoMaPOI's inference, we introduce a \textbf{Reverse Reasoning Fine-Tuning (RRF)} strategy. This strategy generates high-quality backward reasoning samples for each agent, enabling them to learn their specific tasks effectively.

In summary, the main contributions of this paper are as follows:

\begin{itemize}[leftmargin=*,topsep=2pt,itemsep=2pt,parsep=0pt]
    \item \textbf{Multi-agent collaboration for POI prediction.} To the best of our knowledge, this is the first work to introduce an LLM-based multi-agent collaborative mechanism for POI prediction. It systematically explores the effectiveness of agent collaboration, profile construction, candidate enhancement, and reverse reasoning fine-tuning in complex spatiotemporal tasks.
    
    \item \textbf{Converting trajectory into language.} Utilizing our specially designed statistical analysis tools, the Profiler agent transforms numerical check-in data into temporal and spatial statistics, which are then abstracted into linguistic descriptions. This offers a novel way for LLMs to comprehend spatiotemporal data.
    
    \item \textbf{Dynamic constraint on the candidate space.} The Forecaster agent retrieves POIs relevant to the user's current behavior using attribute similarity and re-ranks them based on mobility preferences. This method significantly reduces both the candidate space and prediction error, with its effectiveness validated experimentally and theoretically.
    
    \item \textbf{Extensive experiments and performance gains.} Experiments conducted on three mainstream benchmark datasets demonstrate that the proposed framework significantly outperforms state-of-the-art methods, achieving approximately a 5\%-10\% improvement across all metrics compared to the best baseline.
    
\end{itemize}

The remainder of this paper is organized as follows. Section \ref{section2} reviews related work on traditional POI prediction models and the use of LLMs in POI task, highlighting key challenges. Section \ref{section3} provides definitions for some key concepts and the problem. Section \ref{section4} details the proposed CoMaPOI framework, including its design, key modules (Profiler, Forecaster, and Predictor), and the RRF strategy. Additionally, we theoretically analyze the impact of providing a high-quality candidate set on reducing prediction errors. Section \ref{section5} describes the experimental setup, evaluation metrics, and results, discussing the effectiveness of each module and the overall framework. Finally, Section \ref{section6} concludes the paper and suggests future research directions.

\section{Related Work}
\label{section2}


\paragraph{\textbf{Traditional Models for POI Prediction.}}
Traditional POI prediction models primarily leverage deep learning techniques to capture the high-order dependencies between users and POIs \cite{zeng2025global, zhong2024scfl, yin2023next}. Early approaches often employ sequence models, such as RNNs and LSTMs \cite{huang2019attention}, to enhance the representation of check-in features and improve the understanding of user preferences. SASRec \cite{kang2018self} combines attention mechanisms with RNNs to identify correlations between users and check-in data. BERT4Rec \cite{BERT4Rec} adopts BERT's bidirectional masking strategy to model user check-in sequences and uses the Cloze objective for recommendation. STGN \cite{zhao2020go} focuses on modeling both long-term and short-term preferences within an LSTM-based framework to strengthen the connections between check-ins. TiCoSeRec \cite{dang2023uniform} addresses the challenge of preference drift in sequential data by introducing a series of enhancement operations. These models leverage self-attention or bidirectional Transformer architectures to capture sequential user behaviors, enhancing the modeling of temporal dynamics \cite{zeng2023lgsa}.

To further enhance spatiotemporal information representation, some studies explore contrastive learning, graph-based modeling, and diffusion models \cite{yan2023spatio, yin2023next}. DuoRec \cite{qiu2022contrastive} employs contrastive regularization to reconstruct sequence representation distributions, while DiffuRec \cite{li2023diffurec} applies diffusion models to sequence recommendation to handle item representation and inject uncertainty. Other works, such as GetNext \cite{yang2022getnext} and MTNet \cite{huang2024learning}, integrate multi-task learning or graph-based modeling to enhance multidimensional data representation. POIGDE \cite{yang2024siamese} explicitly utilizes the continuous evolution of user interests by solving the graph differential equation for user interaction behavior. Despite these advancements, these methods often lack sufficient semantic understanding of spatiotemporal data \cite{wang2024embracing}, resulting in performance degradation when handling diverse user behaviors and large-scale POIs.

\paragraph{\textbf{LLM-based Models for Recommendations.}}
To address the limitations in semantic information fusion, recent studies have begun exploring the integration of LLMs in recommendation tasks \cite{bao2023large, zhao2024let}. With their powerful contextual understanding and reasoning capabilities, LLMs can transform user behavior logs and item attributes into natural language representations, improving user preference modeling \cite{tsai2024leveraging, tan2024idgenrec}. For instance, LLMRec \cite{wei2024llmrec} enhances graph data by using LLM-generated outputs to enrich node and edge semantics, improving downstream recommendation performance. A-LLMRec \cite{kim2024large} extracts collaborative knowledge from traditional filtering models to integrate LLM capabilities. Additionally, methods like RDRec \cite{wang2024rdrec} focus on rationale distillation to learn LLM-generated reasoning processes, improving user-item relevance modeling. BAHE \cite{geng2024breaking} employs hierarchical encoding to separate user behavior representations from interactions, improving CTR modeling efficiency. However, LLM predictions heavily rely on having seen all possible items during fine-tuning, making it challenging to handle cold-start items effectively.

\paragraph{\textbf{LLM-based Models for Temporal and Sequential Tasks.}}
In temporal and sequential decision-making domains, prior studies have demonstrated the potential of LLMs in processing time-series data \cite{liu2025st, yuan2024back, qin2025taylors}. By incorporating external retrieval modules, knowledge bases, or structured prompts, LLMs can better understand domain-specific spatiotemporal patterns, enabling future state prediction or complex decision-making \cite{nakshatri2023using, yu2023harnessing}. For instance, TIME-LLM \cite{jin2023time} adapts LLM input and output layers for time-series data, leveraging fine-tuning for temporal predictions. LLMLight \cite{lai2023large} applies LLMs to traffic light decision-making, using refined prompt strategies to significantly reduce average vehicle waiting times. $S^2$IP-LLM \cite{pan2024textbf} aligns temporal and semantic embeddings in joint space for time-series prediction. However, these methods often require structural modifications to LLMs, which diminish the LLM's core language abilities and increase model deployment complexity. Thus, achieving deep spatiotemporal understanding and efficient inference without compromising LLM's core strengths remains a challenge \cite{feng2024move, liu2024can}.

\paragraph{\textbf{LLM-based Models for POI Prediction.}}
POI prediction combines recommendation tasks focusing on user-item (POI) relationships with time-series analysis to capture spatiotemporal behavior patterns \cite{lai2023multi, wang2024dsdrec}. For instance, LLMMove \cite{feng2024move} integrates user geographic preferences, spatial distances, and sequence transitions to frame POI prediction as a ranking problem. LLM4POI \cite{li2024large} uses similar users’ data to assist LLMs in POI prediction, improving performance through consensus information. \citet{kim2024poi} employ LLMs to extract POI-related information from social media, enriching sparse POI datasets through named entity recognition.

Despite these efforts, most approaches focus on data or feature augmentation without addressing the core challenges of applying LLMs to POI prediction. First, LLMs pre-trained on textual data lack the ability to interpret sparse spatiotemporal numerical data, limiting their capacity to model check-in behavior distributions. Second, the absence of effective constraints on large-scale POI candidates often results in uncontrolled generation \cite{chen2024quantifying}, increasing prediction uncertainty and errors. To tackle these challenges, this paper proposes converting users' spatiotemporal information into natural language descriptions and constructing refined POI candidate sets to constrain the candidate space. The design and validation of these methods are detailed in subsequent sections.

\section{Preliminaries}
\label{section3}
In this section, we introduce the core concepts and notations for the POI prediction task.

We define the set of users and the set of POIs as $\mathcal{U} = \{u_1, u_2, \dots, u_N\}$ and $\mathcal{P} = \{p_1, p_2, \dots, p_M\}$, respectively. Each check-in behavior of a user corresponds to a POI. The definitions are as follows:

\textbf{Point of Interest (POI).}
A POI is a specific spatial item associated with a geographical location that users may visit. Each POI $p \in \mathcal{P}$ is represented by a tuple $p = \langle l, \text{lat}, \text{lon}, c \rangle$, where $l$ is the unique identifier of the POI, $\text{lat}$ and $\text{lon}$ are its latitude and longitude, and $c$ denotes its category (e.g., restaurant, park).

\textbf{Check-In.}
A user's check-in behavior is defined as a tuple $x = \langle u, p, t \rangle$, where $u \in \mathcal{U}$ represents the user, $p \in \mathcal{P}$ is the POI visited, and $t \in \mathcal{T}$ is the timestamp. Each check-in constitutes a trajectory point in the user's trajectory.

\textbf{Historical Trajectory.}
The historical trajectory $\mathcal{T}_u^{H}$ includes the user's earlier check-ins, reflecting long-term preferences.
\begin{equation}
\small
\mathcal{T}_u^{H} = \{ \langle p_{u,1}, t_{u,1} \rangle, \dots, \langle p_{u,n_u - L - 1}, t_{u,n_u - L - 1} \rangle \}.
\end{equation}

\textbf{Current Trajectory.}
The current trajectory consists of the user's most recent $L$ check-ins, capturing short-term preferences:
\begin{equation}
\small
\mathcal{T}_u^{C} = \{ \langle p_{u,n_u - L}, t_{u,n_u - L} \rangle, \dots, \langle p_{u,n_u - 1}, t_{u,n_u - 1} \rangle \}.
\end{equation}

\textbf{Candidate Space.}
The candidate space, denoted as $\mathcal{C} \subseteq \mathcal{P}$, is a subset of the complete set of POIs, representing the potential next destinations for a user. In an unconstrained scenario, the candidate space $\mathcal{C}$ is identical to the full set of POIs, $\mathcal{P}$.

\textbf{Problem Definition.} Based on the above definitions, the POI prediction task aims to predict the next POI that a user $u$ is most likely to visit, given their current trajectory $\mathcal{T}_u^{C}$. The historical trajectory $\mathcal{T}_u^{H}$ serves as training data to learn the predictive model. The model generates the predicted POI $\hat{p}_{u,n_u}$ to complete the sequence. Formally, the prediction is defined as $\hat{p}_{u,n_u} = f(\mathcal{T}_u^{C})$, where $f(\cdot)$ represents the predictive function (e.g., implemented via LLMs or traditional POI models).

\section{Method}
\label{section4}

\begin{figure*}[htbp]
\vspace{-10pt}
    \centering
    \includegraphics[width=0.95\textwidth]{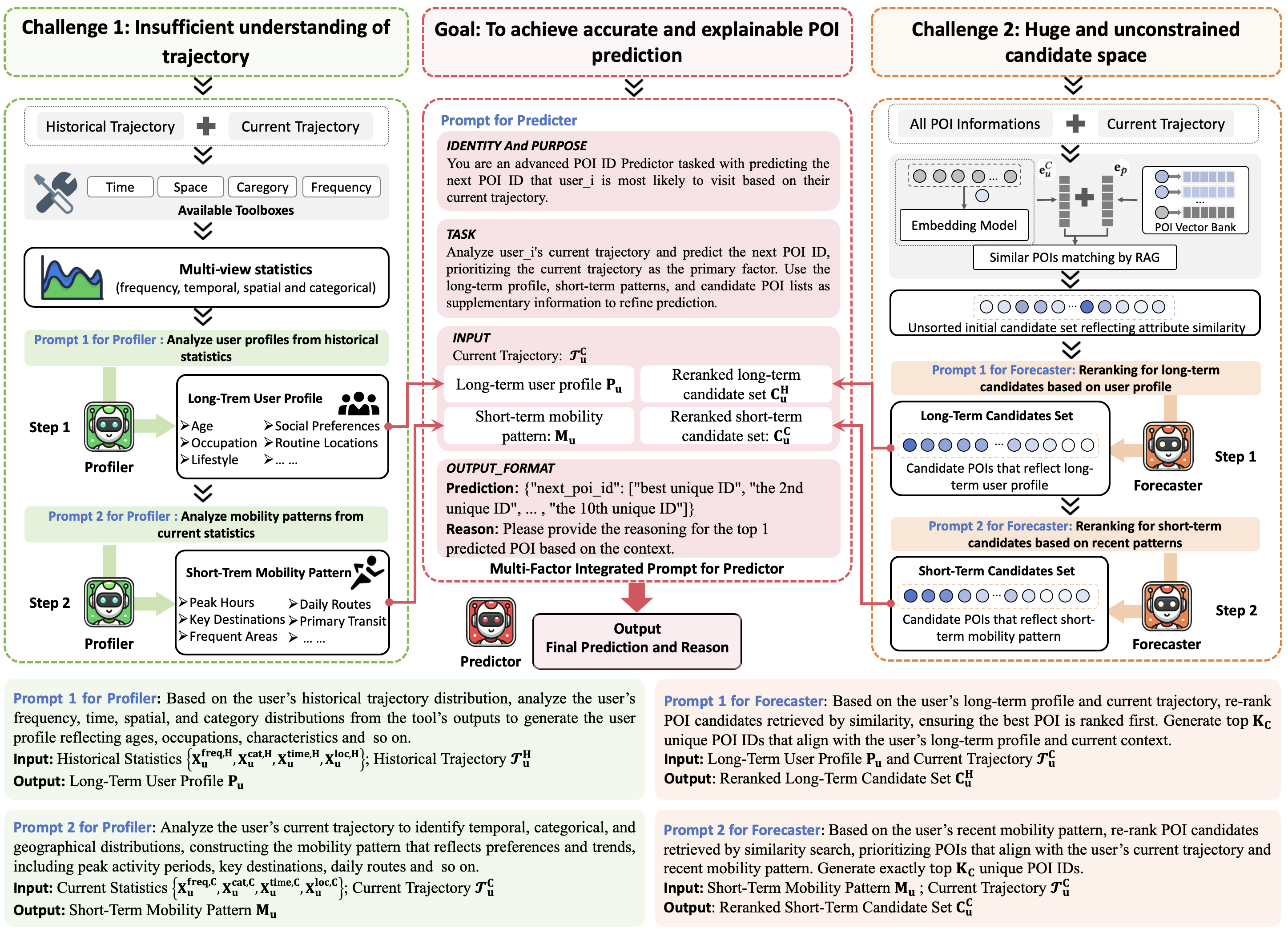} 
    \vspace{-10pt}
    \caption{The overview of the CoMaPOI framework, consisting of three specialized agents (Profiler, Forecaster, and Predictor) collaborating to generate accurate next POI predictions. The process involves extracting long-term user profiles and short-term mobility patterns, retrieving constrained candidate set, and finally producing the prediction.}
    \label{fig_1}
\end{figure*}


To address the challenges of POI prediction with Large Language Models (LLMs), we introduce the \textbf{Collaborative Multi-Agent Framework for Next POI Prediction (CoMaPOI)} (see Figure \ref{fig_1}). This framework consists of three agents: the \textbf{Profiler}, the \textbf{Forecaster}, and the \textbf{Predictor}. The Profiler converts user trajectories into natural language by extracting temporal, spatial, and frequency data, thereby enabling LLMs to better interpret sparse spatiotemporal information. The Forecaster uses retrieval-enhanced techniques to generate and refine a candidate set with insights from user profiles and mobility patterns. The Predictor then integrates these inputs to achieve precise POI predictions. Additionally, CoMaPOI incorporates a Reverse Reasoning Fine-Tuning (RRF) strategy during fine-tuning to reverse-engineer optimal labels, enhancing task-specific performance. Finally, we provide a theoretical analysis of candidate space optimization in the sections that follow.


\subsection{Multi-Agent Collaborative Inference}
\paragraph{\textbf{Profiler: From Trajectory to Language Understanding}}
We observe that due to the discrete nature and high precision of trajectory data, LLMs struggle with accurately extracting visitation frequencies and constructing user preferences from check-in data. Thus, establishing intermediary information that is more regular and comprehensible for LLMs is essential. Inspired by tool-based agent applications \cite{zhuang2023toolqa, zhao2024let}, we develop statistical tools targeting the key aspects of POI prediction. These tools enable the Profiler to extract user profiles and mobility patterns, addressing LLM limitations in processing trajectories.

The toolset includes the time, spatial, category and POI frequency distribution tools. The outputs from these tools describe the distribution of trajectories from a specific perspective in a natural language format. For example, the output of $\text{Tool}_{\text{cat}}$ might be: ``\textit{The historical trajectory of User 27 shows the following category distribution: Home (private), Frequency: 117; Government Building, Frequency: 51; Gym / Fitness Center, Frequency: 25; …; Airport, Frequency: 2; Sporting Goods Shop, Frequency: 1.}'' The complete toolbox can be represented as $\text{Tools} = \{\text{Tool}_{\text{freq}}, \text{Tool}_{\text{cat}}, \text{Tool}_{\text{time}}, \text{Tool}_{\text{loc}}\}$.

The user’s long-term trajectory, while temporally distant from the next check-in and less reflective of recent preferences, offers valuable insights into macro-level characteristics. Accordingly, the long-term prompt emphasizes constructing a comprehensive user profile, including attributes such as age, occupation, and social preferences. To achieve this, the Profiler leverages tools to analyze and extract statistical descriptions from the user's historical trajectory: 
\begin{equation}
\small
\{X_u^{\text{freq},H}, X_u^{\text{cat},H}, X_u^{\text{time},H}, X_u^{\text{loc},H}\} = \text{Tools}(\mathcal{T}_u^{H}),
\end{equation}
where $\{X_u^{\text{freq},H}, X_u^{\text{cat},H}, X_u^{\text{time},H}, X_u^{\text{loc},H}\}$ represent the frequency distributions for POI frequency, category, time, and space, respectively. The Profiler then integrates these statistical results into the user's long-term profile through the LLM, as shown in \textit{Prompt 1 for Profiler} in Figure \ref{fig_1}, constructing the user's long-term profile $P_u$:
\begin{equation}
\small
P_u \leftarrow \text{LLM}(\{X_u^{\text{freq},H}, X_u^{\text{cat},H}, X_u^{\text{time},H}, X_u^{\text{loc},H}\}, \mathcal{T}_u^{H}).
\end{equation}

For specific examples of user profiles $P_u$, see the first part of the Profiler's output in the case study section in Figure~\ref{fig_case}.

Moreover, the Profiler also needs to perform the same operations on the current trajectory. For the current trajectory, we consider it more aligned with target check-ins and therefore more focused on specific preferences. Thus, the prompt for the current trajectory emphasizes analyzing the user's recent mobility patterns, like peak hours, key destinations, daily routes and so on. The output of mobility patterns can be expressed with the following formula: 
\begin{equation} 
\small
M_u \leftarrow \text{LLM}(\{X_u^{\text{freq},C}, X_u^{\text{cat},C}, X_u^{\text{time},C}, X_u^{\text{loc},C}\}, \mathcal{T}_u^{C}), \end{equation}
where the statistical information of the current trajectory is obtained by $\text{Tools}(\mathcal{T}_u^{C})$. The prompts for generating mobility patterns $M_u$ can be referenced in \textit{Prompt 2 for Profiler} in Figure \ref{fig_1}. For specific examples of mobility patterns $M_u$, see the second part of the Profiler's output in the case study section in Figure~\ref{fig_case}.

Through the above process, the Profiler leverages tool functions to transform discrete numerical trajectories into more structured statistical information, further refining the user's long-term profiles and recent mobility patterns. These language descriptions provide richer context for the subsequent agents, enhancing their performance.

\paragraph{\textbf{Forecaster: Reducing the candidate Space}}
The Forecaster primarily addresses the challenge of narrowing the candidate space for prediction. In POI prediction tasks, a city can have hundreds of thousands of POIs, making it impractical to provide all POI information to the LLM. Therefore, creating a high-hit-rate candidate set is a feasible alternative. Inspired by Retrieval-Augmented Generation (RAG) \cite{cuconasu2024power, lyu2024crud}, we develop a RAG strategy tailored for POI scenarios, using semantic similarity retrieval to return a set of candidate POIs. 

To generate an initial candidate set, we convert all POI information into embedding vectors stored in a vector database. During inference, we generate embedding vectors for the current trajectory's POIs and match them with the most relevant POIs in the database. These similar POIs form the initial candidate set. Each POI is converted into a check-in description, e.g., \textit{"POI $p$ is located at coordinates ($lat$, $lon$), category: $c$"}. This description is then transformed into a semantic vector tuple $(p, \mathbf{e}_p)$ using the BCE embedding model \cite{youdao_bcembedding_2023}, where $p$ is the POI ID and $\mathbf{e}_p$ is the corresponding embedding. All tuples are stored in an embedding database $V = \{(p, \mathbf{e}_p) \mid p \in \mathcal{P}\}$.

During the prediction phase, the Forecaster queries $V$ by computing an embedding $\mathbf{e}_{p_{u,i}}$ for each POI $p_{u,i}$ in $\mathcal{T}_u^{C}$ and retrieving the top-$K_C$ most similar POIs based on:
\begin{equation}
\small
C_{p_{u,i}} = \mathop{\arg\max}\limits_{C \subseteq \mathcal{P}, |C|=K_C} \sum_{p \in C} \text{sim}(\mathbf{e}_{p_{u,i}}, \mathbf{e}_p),
\end{equation}
where $\text{sim}(\mathbf{e}_{p_{u,i}}, \mathbf{e}_p) = \frac{\mathbf{e}_{p_{u,i}} \cdot \mathbf{e}_p}{\|\mathbf{e}_{p_{u,i}}\| \|\mathbf{e}_p\|}$. Aggregating over all $p_{u,i} \in \mathcal{T}_u^{C}$ yields the initial candidate set $C_u^{C,\text{init}} = \bigcup_{p_{u,i} \in \mathcal{T}_u^{C}} C_{p_{u,i}}$.

To address the insufficient hit rate of $C_u^{C,\text{init}}$ observed in subsequent evaluations (see Table~\ref{tab_2}), we introduced the Forecaster agent to optimize the candidate set. It mitigates the impact of noise by ensuring the label is ranked higher and reordering candidates based on user profiles and mobility patterns from the Profiler, effectively prioritizing high-probability POIs and reducing the candidate size (e.g., from 250 to 25).

From a long-term perspective, we prompt the Forecaster to optimize $C_u^{C,\text{init}}$ based on the user profile $P_u$, involving reordering and narrowing down the number of candidates. Specific prompts can be found in \textit{Prompt 1 for Forecaster} in Figure \ref{fig_1}, resulting in the long-term candidate set:

\begin{equation}
\small
C_u^H \leftarrow \text{LLM}(P_u, C_u^{C,\text{init}}).
\end{equation}

Simultaneously, as shown in \textit{Prompt 2 for Forecaster} in Figure \ref{fig_1}, the Forecaster also optimizes $C_u^{C,\text{init}}$ based on the recent mobility pattern $M_u$ to generate the short-term candidate set:
\begin{equation}
\small
C_u^C \leftarrow \text{LLM}(M_u, C_u^{C,\text{init}}).
\end{equation}

Finally, the Forecaster outputs $C_u^H$ and $C_u^C$, representing refined candidate sets aligned with the user's long-term interests and short-term behavioral characteristics, respectively. These high-quality candidate sets provide valuable input for the Predictor, enabling more accurate prediction decisions.

\paragraph{\textbf{Predictor: Integrated Decision-Making}}

As shown in the prompt (\textit{Prompt for Predictor}) in Figure \ref{fig_1}, the Predictor's prediction prompt has been expanded beyond the recent trajectory $\mathcal{T}_u^{C}$ to include the long-term user profiles $P_u$, short-term patterns $M_u$, long-term candidates $C_u^{H}$, short-term candidates $C_u^{C}$, as well as the recent trajectory $\mathcal{T}_u^{C}$. Here, $P_u$ and $M_u$ assist the Predictor in understanding the deeper user mobility intentions behind the discrete trajectories, while $C_u^{H}$ and $C_u^{C}$ provide a more focused candidate space for prediction. Based on these information, the Predictor's prediction is represented as:
\begin{equation}
\small
\hat{p}_{u,n_u} \leftarrow \text{LLM}(\{P_u, M_u, C_u^{H}, C_u^{C}\}, \mathcal{T}_u^{C}),
\end{equation}
where the Predictor can select a POI from $C_u^{H} \cup C_u^{C}$ or, if necessary, opt for a POI outside this candidate set. This integration ensures that the Predictor effectively balances long-term user preferences, short-term mobility insights, and the flexibility to explore candidate options, leading to a robust and context-aware decision-making process.

\subsection{Reverse Reasoning Fine-Tuning (RRF)}
To further enhance the accuracy of CoMaPOI predictions, we need to fine-tune each agent based on task-specific requirements. A prominent issue is the lack of labels corresponding to CoMaPOI sub-tasks in the POI data. For instance, the Profiler lacks ideal user profiles and mobility patterns as labels, while the Forecaster lacks ideal candidate sets. Although the Predictor has labels, its input prompts lack relevant content. Therefore, before fine-tuning, we need to reverse-engineer the required labels or prompts.

To address this issue, we propose a method called Reverse Reasoning Fine-Tuning (RRF) to supplement the missing labels by reversing the inference process. Simply put, by providing the correct POI ID $p_{u,n_u}$, the agents reverse-engineer the corresponding user profiles, mobility patterns, and candidate sets. During this process, ensure that the output can guide the LLM to accurately determine the correct $p_{u,n_u}$. For example, the candidate set must include $p_{u,n_u}$, and the mobility pattern descriptions must be highly relevant to $p_{u,n_u}$. The reverse reasoning process of RRF is essentially about creating labels that maximize the probability of correctly predicting the true POI:
\begin{align}
\small
(P_u^{+}, M_u^{+}) &\leftarrow \arg\max_{P, M} S(P_u, C_u^{H}; p_{u,n_u}), \\
(C_u^{H+}, C_u^{C+}) &\leftarrow \arg\max_{M, C} S(M_u, C_u^{C}; p_{u,n_u}),
\end{align}
where, $S(\cdot)$ denotes alignment scores that measure the association between the constructed content and $p_{u,n_u}$.

Once the reverse reasoning is complete, we obtain the ideal user profiles, mobility patterns, and candidate sets, which serve as labels for fine-tuning the Profiler and Forecaster. For the Predictor, its task is to predict the next POI by integrating the current trajectory $\mathcal{T}_u^C$ with the additional information. $(P_u^{+}, M_u^{+}, C_u^{H+}, C_u^{C+})$ serve as additional information in the prompts, constructing the fine-tuning samples for the Predictor.

After reverse reasoning, the fine-tuning samples required for CoMaPOI are constructed. The next step involves fine-tuning the LLM, where the optimization goal is to integrate the three agents (Profiler, Forecaster, and Predictor) into a unified optimization objective:
\begin{equation}
\small
(\hat{\theta}_p, \hat{\theta}_f, \hat{\theta}_r) = \mathop{\arg\min}\limits_{\theta_p, \theta_f, \theta_r} \sum -\log P_{\theta}(p_{u,n_u} \mid X^{+}),
\end{equation}
where $X^{+}$ represents the backward-constructed reasoning samples, including $(P_u^{+}, M_u^{+}, C_u^{H+}, C_u^{C+})$ and all relevant prompts.

During the fine-tuning process, we adopt LoRA \cite{lora2021} and quantization techniques \cite{unsloth} to accelerate computation and reduce costs. This process is implemented using the TRL framework \cite{vonwerra2022trl}, which ensures efficiency and scalability. After fine-tuning with RRF, the model is capable of producing forward reasoning paths during inference that are more likely to align with the true causal relationships.

\subsection{Theoretical Analysis of Candidate Space Reduction}
\label{sec_4.4}
When no candidate POI is provided, LLM's prediction of the POI can be viewed as identifying the next POI over the entire range of POIs, which we call a global search. The vast and unconstrained range of candidates in global retrieval can increase prediction errors. To address this, in CoMaPOI, we provide an optimized candidate POI set to reduce the candidate space, thus lowering prediction errors. Naturally, a key question arises: \textit{"How do we determine that providing a candidate POI set indeed reduces prediction error compared to global search, rather than introducing more noise?"}

Let the set of all possible POIs be $\mathcal{P}$ of size $M$, and let $p^*$ be the true label for the user. In the global search, the posterior probability of LLM correctly predicting can be defined as $P\bigl(p^*\mid \mathcal{T}_u^C\bigr)$. Consequently, the probability of failing to predict $p^*$ under global search is represented as:
\begin{equation}
\small
Error_{\mathrm{global}} = 1 - P\bigl(p^*\mid \mathcal{T}_u^C\bigr).
\end{equation}

In CoMaPOI, the prediction probability is calculated based on the optimized candidate set. Suppose the optimized candidate set is $C \subseteq \mathcal{P}$, where $|C| \ll M$. Whether the candidate set itself contains the key for label $p^*$ is represented by $\mathcal{P}(p^* \in C)$, which can also be considered as the hit rate of the candidate set for the label. 

After providing a candidate set, the prediction error comes from two scenarios: the first is that the candidate set contains the label, but the LLM makes an incorrect prediction, represented by $\mathcal{P}(p^* \in C)\bigl[1 - P\bigl(p^*\mid \mathcal{T}_u^C, p^* \in C\bigr)\bigr]$. The second scenario is that the candidate set does not contain the label, and the LLM still makes an incorrect prediction, represented by $\bigl[1 - \mathcal{P}(p^* \in C)\bigr]\bigl[1 - P\bigl(p^*\mid \mathcal{T}_u^C, p^* \notin C\bigr)\bigr]$. Therefore, in CoMaPOI, the prediction error can be expressed as:
\begin{equation}
\small
\begin{aligned}
Error_{\mathrm{with\,candidate}} & = \mathcal{P}(p^* \in C)\bigl[1 - P\bigl(p^*\mid \mathcal{T}_u^C, p^* \in C\bigr)\bigr] \\
& \quad + \bigl[1 - \mathcal{P}(p^* \in C)\bigr]\bigl[1 - P\bigl(p^*\mid \mathcal{T}_u^C, p^* \notin C\bigr)\bigr].
\end{aligned}
\end{equation}

A necessary condition for the candidate-based approach to outperform the global search can be expressed as:
\begin{equation}
\small
Error_{\mathrm{with\,candidate}} < Error_{\mathrm{global}}.
\end{equation}

By substituting the above definitions and rearranging, we obtain:
\begin{equation}
\small
\begin{aligned}
& \mathcal{P}(p^* \in C)P\bigl(p^*\mid \mathcal{T}_u^C, p^* \in C\bigr) \\
& \quad + \bigl(1 - \mathcal{P}(p^* \in C)\bigr)P\bigl(p^*\mid \mathcal{T}_u^C, p^* \notin C\bigr) > P\bigl(p^*\mid \mathcal{T}_u^C\bigr),
\end{aligned}
\end{equation}

Then isolating $\mathcal{P}(p^* \in C)$, we obtain:
\begin{equation}
\small
\mathcal{P}(p^* \in C) > \frac{P\bigl(p^*\mid \mathcal{T}_u^C\bigr) - P\bigl(p^*\mid \mathcal{T}_u^C, p^* \notin C\bigr)}
{P\bigl(p^*\mid \mathcal{T}_u^C, p^* \in C\bigr) - P\bigl(p^*\mid \mathcal{T}_u^C, p^* \notin C\bigr)}.
\label{eq_17}
\end{equation}

Equation \ref{eq_17} indicates that when the hit rate of the candidate set $\mathcal{P}(p^* \in C)$ is higher than the right side (considered as a lower bound), it ensures that the prediction error based on the candidates is lower than the prediction error of the global search. The optimization strategy in CoMaPOI promotes an increase in the hit rate of the candidate set, i.e., it increases $\mathcal{P}(p^* \in C)$. Additionally, it enhances the probability of successful predictions based on the candidates, i.e., $P\bigl(p^*\mid \mathcal{T}_u^C, p^* \in C\bigr)$ is increased. This theoretically proves that the method of providing high-quality candidate set in CoMaPOI effectively reduces prediction error. Moreover, Table \ref{tab_2} in this paper demonstrates that the candidate set hit rates under CoMaPOI on three datasets satisfy the inequality of eq~\ref{eq_17}. For example, CoMaPOI in NYC, $\mathcal{P}(p^* \in C)$ reached 67.51\%, and $P\bigl(p^*\mid \mathcal{T}_u^C, p^* \in C\bigr)$ reached 88.16\%. The lower bound was calculated to be 45.27\%, satisfying the inequality $67.51\% > 45.27\%$.

\section{ExperimentsExperiments}
\label{section5}

In this section, we comprehensively evaluate the proposed \textbf{CoMaPOI} framework for POI prediction through extensive experiments. The evaluation emphasizes comparisons between CoMaPOI and state-of-the-art (SOTA) baselines to demonstrate its effectiveness. Next, we analyze CoMaPOI's scalability across various LLM architectures, including smaller parameter LLMs, underscoring its applicability in resource-constrained scenarios. Additionally, we conduct ablation studies to examine performance variations when different modules are removed. Furthermore, we analyze the effects of candidate space optimization and visualization, as well as the impact of hyperparameter tuning for the candidate set size. Finally, in the case study, we present CoMaPOI’s intermediate outputs during reasoning and its ability to produce accurate and interpretable predictions. The following subsections detail the experimental setup, results, and analysis.

\subsection{Experimental Setting}

\paragraph{\textbf{Datasets.}}
We use three publicly available POI datasets: \textbf{NYC}, \textbf{TKY}, and \textbf{CA}. The NYC dataset contains check-ins collected in New York City from April 2012 to February 2013, while the TKY dataset includes check-ins from Tokyo during the same time period \cite{yang2014modeling}. The CA dataset consists of check-ins from California and Nevada, collected between February 2009 and October 2010 \cite{yuan2013time}. Each record in these datasets contains a user ID, POI ID, POI category, GPS coordinates, and a timestamp. Consistent with previous studies \cite{yang2022getnext, huang2024learning}, we excluded users and POIs with a low number of check-ins. After preprocessing, the NYC dataset contains 988 users, 5086 POIs, and 99,964 check-ins; the TKY dataset contains 2206 users, 7849 POIs, and 325,313 check-ins; the CA dataset contains 1818 users, 13,564 POIs, and 174,791 check-ins. For each user in these datasets, we designate the last 30 visited points as the test set, also referred to as the current trajectory $\mathcal{T}_u^{C}$, while all preceding points constitute the training set, referred to as the historical trajectory $\mathcal{T}_u^{H}$. The fixed trajectory length in the test set is designed to control the token length input to the LLM, ensuring a consistent token budget that effectively prevents memory overflow.

\paragraph{\textbf{Baseline Methods.}}
We compare CoMaPOI with 11 state-of-the-art methods, which can be categorized into four groups. Sequential Models, such as \textbf{SASRec} \cite{kang2018self}, \textbf{BERT4Rec} \cite{BERT4Rec}, and \textbf{TiCoSeRec} \cite{dang2023uniform}, leverage Transformer architectures to directly model user check-in sequences, improving recommendation accuracy. Graph-based or Contrastive methods, including \textbf{GETNext} \cite{yang2022getnext}, \textbf{CrossDR} \cite{tao2023next}, \textbf{MAERec} \cite{ye2023graph}, and \textbf{DuoRec} \cite{qiu2022contrastive}, incorporate graph neural networks, contrastive learning, and spatio-temporal information to address data sparsity and cold start challenges, enhancing recommendation robustness. Generative and Diffusion-based methods, such as \textbf{POIGDE} \cite{yang2024siamese}, \textbf{MTNet} \cite{huang2024learning}, and \textbf{DiffuRec} \cite{li2023diffurec}, employ generative techniques or diffusion frameworks to capture the distribution of potential next POIs, providing diverse and accurate predictions. Lastly, LLM-based methods, like \textbf{LLM4POI} \cite{li2024large}, introduce LLM into the POI prediction pipeline, representing a baseline for LLM-enhanced solutions.

\paragraph{\textbf{Evaluation Metrics.}}
We assess model performance using standard ranking-based metrics commonly used in POI prediction tasks: \textbf{HR@k}, indicating the proportion of test cases where the correct POI appears in the top $k$ predictions; \textbf{NDCG@k}, reflecting both accuracy and the ranking quality of predictions; and \textbf{MRR}, which evaluates the position of the correct answer by calculating the reciprocal of its rank.

\paragraph{\textbf{Implementation Details.}}
During CoMaPOI’s inference, we set top\_p to 1, temperature to 0, and n to 1 to eliminate sampling randomness. For the fine-tuning of CoMaPOI, we adopt a learning rate of $1\text{e}^{-4}$, a batch size of 16, and a maximum of 200 steps. Additionally, we used 50 warm-up steps and applied a weight decay of 0.01 to balance computational efficiency and optimization stability. Both inference and fine-tuning of CoMaPOI were conducted on three NVIDIA H800 80G GPUs. The foundational LLM for CoMaPOI is Llama3.1:8b \cite{sreenivas2024llm}. Additionally, in the application validation experiments, we also tested other LLMs, including Qwen2.5 \cite{qwen} and Mistral \cite{jiang2023mistral7b}.

\subsection{Surpassing SOTA Baselines}
\label{sec:q1_analysis}
\begin{table*}[htpb]
    \centering
    \caption{Performance Comparison Between CoMaPOI and Baselines Across Three Datasets}
    \vspace{-10pt}
    \resizebox{\textwidth}{!}{ 
    \begin{tabular}{lccccccccccccccc}
        \toprule
        \multicolumn{1}{c}{\multirow{2}[2]{*}{\textbf{Model}}} & \multicolumn{5}{c}{\textbf{NYC (\%)}} & \multicolumn{5}{c}{\textbf{TKY (\%)}} & \multicolumn{5}{c}{\textbf{CA (\%)}} \\ 
        \cmidrule(lr){2-6} \cmidrule(lr){7-11} \cmidrule(lr){12-16}
        & HR@5 & HR@10 & NDCG@5 & NDCG@10 & MRR 
        & HR@5 & HR@10 & NDCG@5 & NDCG@10 & MRR 
        & HR@5 & HR@10 & NDCG@5 & NDCG@10 & MRR \\ 
        \midrule
        SASRec & 40.38 & 48.28 & 28.98 & 31.26 & 27.01 & 34.72 & 43.25 & 25.48 & 28.37 & 24.70 & 21.34 & 26.40 & 16.04 & 17.68 & 15.94 \\ 
        BERT4Rec & 39.60 & 46.70 & 29.03 & 30.94 & 27.11 & 36.17 & 43.47 & 26.70 & 29.08 & 25.39 & 23.22 & 28.22 & 17.26 & 18.66 & 16.61 \\ 
        DuoRec & 34.72 & 40.69 & 24.54 & 26.34 & 22.76 & 24.66 & 31.19 & 18.77 & 20.89 & 18.67 & 19.47 & 24.04 & 14.47 & 15.85 & 14.31 \\ 
        TiCoSeRec & 40.69 & 48.08 & 29.65 & 32.04 & 27.54 & 29.33 & 37.04 & 21.83 & 24.21 & 21.51 & 22.66 & 28.11 & 16.99 & 18.48 & 16.75 \\ 
        CrossDR & 40.80 & 47.20 & 30.67 & 32.59 & 28.66 & 32.51 & 40.57 & 23.58 & 26.19 & 22.65 & 22.11 & 28.61 & 16.11 & 18.18 & 15.79 \\ 
        MAERec & 37.65 & 44.53 & 27.72 & 29.85 & 26.10 & 32.73 & 40.66 & 23.93 & 26.14 & 23.05 & 24.37 & 29.37 & 17.76 & 19.48 & 17.09 \\ 
        GETNext & 43.60 & 51.30 & 32.56 & 35.13 & 30.97 & 41.11 & 48.50 & 30.75 & 33.21 & 29.21 & 24.26 & 29.32 & 17.83 & 19.48 & 17.16 \\ 
        POIGDE & 34.11 & 38.36 & 26.64 & 27.97 & 25.40 & 28.20 & 31.82 & 22.46 & 23.57 & 21.80 & 22.61 & 26.73 & 18.24 & 19.49 & 17.89 \\ 
        MTNet & \underline{47.57} & \underline{53.24} & 35.85 & 37.69 & 33.40 & \underline{44.15} & \underline{51.22} & \underline{32.99} & \underline{35.37} & \underline{31.18} & \underline{30.25} & \underline{36.08} & \underline{22.39} & \underline{24.27} & \underline{21.35} \\ 
        DiffuRec & 37.25 & 40.79 & 29.16 & 30.23 & 27.74 & 34.50 & 39.89 & 26.89 & 28.70 & 25.94 & 21.67 & 24.15 & 17.35 & 18.06 & 16.60 \\ 
        LLM4POI & 44.64 & 46.66 & \underline{37.49} & \underline{38.16} & \underline{35.90} & 34.90 & 39.48 & 27.23 & 28.22 & 29.72 & 23.65 & 25.91 & 19.01 & 19.39 & 20.13 \\ 
        \midrule
        CoMaPOI (ours) & \textbf{51.62} & \textbf{59.01} & \textbf{40.42} & \textbf{42.82} & \textbf{37.67} & \textbf{45.83} & \textbf{54.26} & \textbf{34.48} & \textbf{37.20} & \textbf{31.82} & \textbf{33.00} & \textbf{39.16} & \textbf{24.96} & \textbf{26.96} & \textbf{23.10} \\ 
        \rowcolor[gray]{0.95} 
        Improvement & \textcolor{black}{+}8.51\% & \textcolor{black}{+}10.84\% & \textcolor{black}{+}7.82\% & \textcolor{black}{+}12.21\% & \textcolor{black}{+}4.93\% & \textcolor{black}{+}3.81\% & \textcolor{black}{+}5.94\% & \textcolor{black}{+}4.52\% & \textcolor{black}{+}5.17\% & \textcolor{black}{+}2.05\% & \textcolor{black}{+}9.09\% & \textcolor{black}{+}8.54\% & \textcolor{black}{+}11.48\% & \textcolor{black}{+}11.08\% & \textcolor{black}{+}8.20\% \\ 
        \bottomrule
    \end{tabular}}
    \label{tab_1}

\end{table*}

To evaluate whether CoMaPOI outperforms SOTA methods, we compare its performance against multiple baselines on the NYC, TKY, and CA datasets, as shown in Table~\ref{tab_1}. Based on LLama3.1:8b, CoMaPOI consistently outperforms all baselines across the three datasets. Specifically, it achieves average improvements over the best performance among the baselines (primarily contributed by MTNet and LLM4POI) by 8.86\% on NYC, 4.3\% on TKY, and 9.68\% on CA across five key metrics.

The results indicate that different baselines demonstrate their own strengths across datasets. MTNet performs well on the NYC and TKY datasets due to its robust sequence modeling capabilities but struggles on the CA dataset, which features a larger geographical scope and more diverse POIs. LLM4POI exhibits competitive performance on metrics such as NDCG, leveraging LLM-based contextual information to effectively capture similarities between users. Other baselines, such as SASRec, POIGDE, and GETNext, incorporate various optimizations but still fall short in addressing the complex spatiotemporal dependencies and diverse user preferences required for accurate predictions. 

In contrast, by leveraging a multi-agent collaborative mechanism, CoMaPOI effectively models both macro-level user profiles and micro-level mobility patterns. Furthermore, its well-designed candidate set filtering narrows the candidate space, reducing prediction errors. These innovations collectively contribute to CoMaPOI’s substantial performance improvements.

\subsection{Generality and Scalability of CoMaPOI}

\begin{figure}[t]
    \vspace{-5pt}

	\centering
	\subfloat[Performance Across Different LLMs Backbones]{%
		\includegraphics[width=1\linewidth]{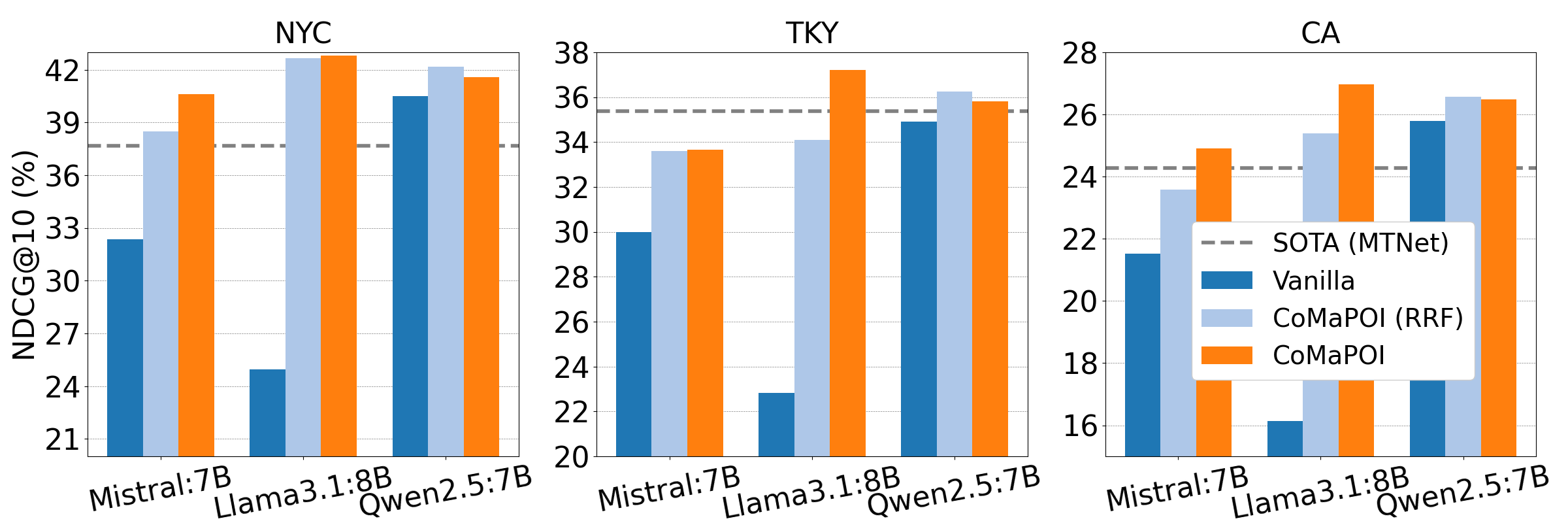}%
		\label{fig:different_llms}%
	}
	\hfill
	\subfloat[Performance on Smaller Parameter LLMs]{%
		\includegraphics[width=1\linewidth]{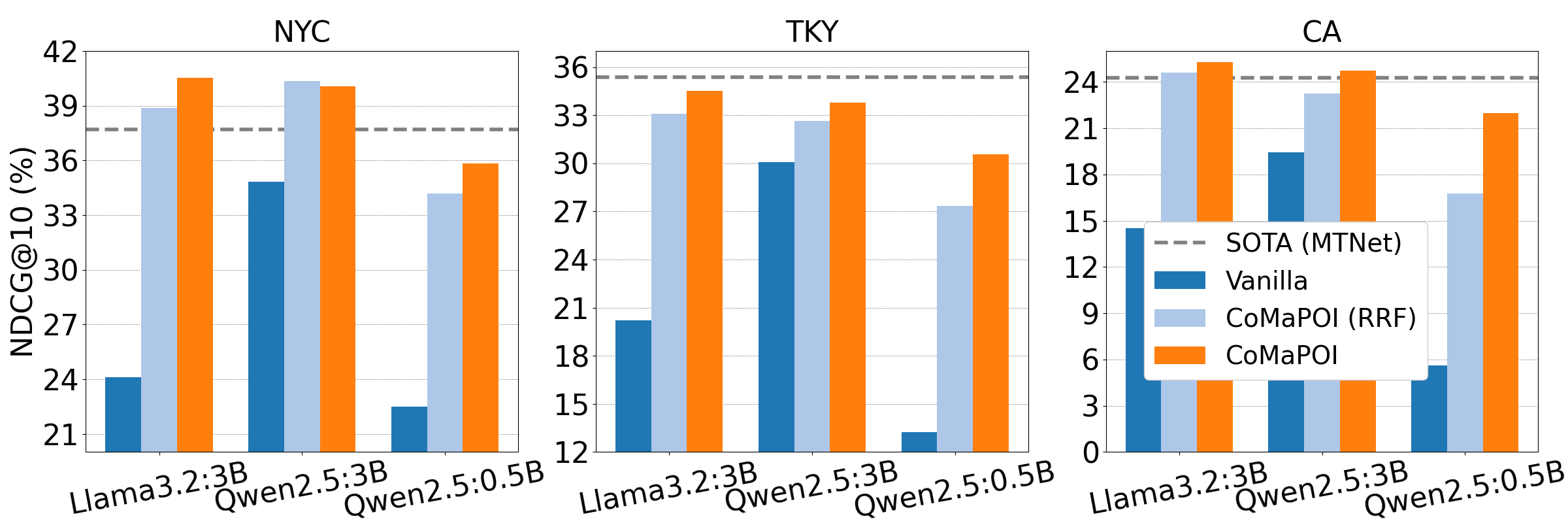}%
		\label{fig:smaller_llms}%
	}
\vspace{-10pt}
	\caption{Performance Evaluation of CoMaPOI Across Different LLMs Architectures and Smaller Parameter LLMs.}
	\label{fig:llm_scalability}
\vspace{-15pt}
\end{figure}

We further analyze the applicability and scalability of CoMaPOI, as shown in Figure~\ref{fig:llm_scalability}. Due to its non-intrusive design that avoids modifications to the LLM architecture, CoMaPOI can be seamlessly applied to various LLM architectures. We evaluate CoMaPOI on several mainstream open-source LLMs, including Mistral:7B, Llama3.1:8B, and Qwen2.5:7B. The experimental results in Figure~\ref{fig:llm_scalability} (a) demonstrate that CoMaPOI significantly enhances POI prediction performance across all three LLMs. The largest improvement was observed on Llama3.1:8B, followed by Mistral:7B. 

In resource-constrained environments, such as mobile devices or in-vehicle systems, POI prediction often requires deploying smaller LLMs. To explore this, we further test CoMaPOI on smaller models in Figure~\ref{fig:llm_scalability} (b), such as Llama3.2:3B and Qwen2.5:0.5B. These smaller models, which generally perform poorly in their original (Vanilla) configurations, exhibit significant improvements when integrated with the CoMaPOI framework. For instance, Llama3.2:3B outperforms the MTNet after applying CoMaPOI.

Additionally, we introduce a lightweight variant of CoMaPOI, called CoMaPOI (RRF). CoMaPOI (RRF) involves using a single agent to perform all tasks in CoMaPOI, where RRF fine-tunes the agent to simultaneously learn the three subtasks: profile construction, candidate optimization, and prediction. Notably, the inference cost under the RRF setting remains comparable to that of directly using a vanilla LLM for reasoning. Although CoMaPOI (RRF) achieves slightly lower performance than the full CoMaPOI framework, it still delivers substantial improvements over vanilla LLMs and outperforms the SOTA. These results highlight the practicality and versatility of CoMaPOI in real-world applications.

\label{sec:q2_analysis}

\subsection{Ablation Study}
\label{sec:q5_analysis}
Table~\ref{tab_abalation} demonstrates the unique contribution of each module to the CoMaPOI framework's performance. \textit{"w/o Profiler"} results in the loss of essential contextual information, causing consistent performance declines across all datasets. \textit{"w/o Forecaster"} leaves the candidate space unconstrained, increasing prediction uncertainty and leading to degraded performance. \textit{"w/o RRF"} significantly reduces prediction accuracy and ranking performance, particularly on the CA dataset, highlighting the critical role of RRF in enhancing agents' understanding and adaptability to user behavior. In the \textit{"w/o Refine"} setting, candidates are directly retrieved from RAG without further optimization, resulting in a broader and less focused candidate space that negatively impacts prediction quality. The \textit{"only Single Agent"} setting lacks the complementary perspectives offered by multi-agent collaboration, leading to diminished results. Finally, the \textit{"only SFT"} baseline relies solely on supervised fine-tuning, showing the weakest performance due to the absence of all modules. These findings underscore the importance of each module, particularly RRF and candidate space optimization, in achieving superior POI prediction performance.

\begin{table}[t]
  \centering
  \Huge
  \caption{Ablation Study of CoMaPOI Framework and the Impact of Removing Key Modules}
    \resizebox{0.48\textwidth}{!}{ 
    \begin{tabular}{lcccccc}
    \toprule
    \multirow{2}[2]{*}{\textbf{Setting}} & \multicolumn{2}{c}{\textbf{NYC (\%)}} & \multicolumn{2}{c}{\textbf{TKY (\%)}} & \multicolumn{2}{c}{\textbf{CA (\%)}} \\
          & HR@10 & NDCG@10 & HR@10 & NDCG@10 & HR@10 & NDCG@10 \\
    \midrule
    CoMaPOI  & 59.01  & 42.82  & 54.26  & 37.20  & 39.16  & 26.96  \\
    w/o Profiler & 58.60  & 40.59  & 52.40  & 34.93  & 32.01  & 22.31  \\
    w/o Forecaster & 56.48  & 39.61  & 49.86  & 34.86  & 35.64  & 23.61  \\
    w/o RRF & 54.35  & 38.40  & 46.51  & 31.83  & 32.84  & 22.86  \\
    w/o Refine & 57.79  & 40.63  & 51.86  & 35.11  & 37.46  & 26.09  \\
    only Single Agent & 56.91  & 40.52  & 51.72  & 35.61  & 37.27  & 25.48  \\
    only SFT & 47.77  & 35.32  & 41.75  & 30.39  & 29.10  & 22.38  \\

    \bottomrule
    \end{tabular}%
    }
  \label{tab_abalation}%

\end{table}

\subsection{Further Study}
\paragraph{\textbf{Analysis of Candidate Space}}

\label{sec:q4_analysis}

Table~\ref{tab_2} presents the candidate set hit rates and related prediction statistics of the CoMaPOI framework across three datasets, serving as validation for the theoretical analysis in Section~\ref{sec_4.4}. A successful hit is defined as the inclusion of the label within the candidate set. From the candidate set stages, the RAG initial candidate set shows a relatively low hit rate (e.g., 43.52\% for NYC). However, with optimizations based on user profiles and movement patterns, the candidate set hit rate improves significantly, reaching 67.51\% for the merged candidate set in the NYC dataset. From the prediction statistics, the posterior probability of the global search without candidate sets is quite low (e.g., 36.03\% for NYC), while the posterior success rate based on the optimized merged candidate set is significantly higher (88.16\% for NYC). This demonstrates that the optimization strategy greatly enhances prediction accuracy. Meanwhile, the out-of-candidate success rate $P(p^* \mid \mathcal{T}_u^C, p^* \notin C)$ varies across datasets, with NYC at $0.31\%$ and TKY at $13.17\%$, reflecting differences in dataset complexity and the robustness of the model's out-of-candidate predictions. Through these results, we successfully verified that CoMaPOI satisfies eq~\ref{eq_17} across all three datasets. 

\begin{figure}[t]

    \centering
    \vspace{0pt}
    \begin{subfigure}[b]{0.2\textwidth}
        \centering
        \includegraphics[width=\textwidth]{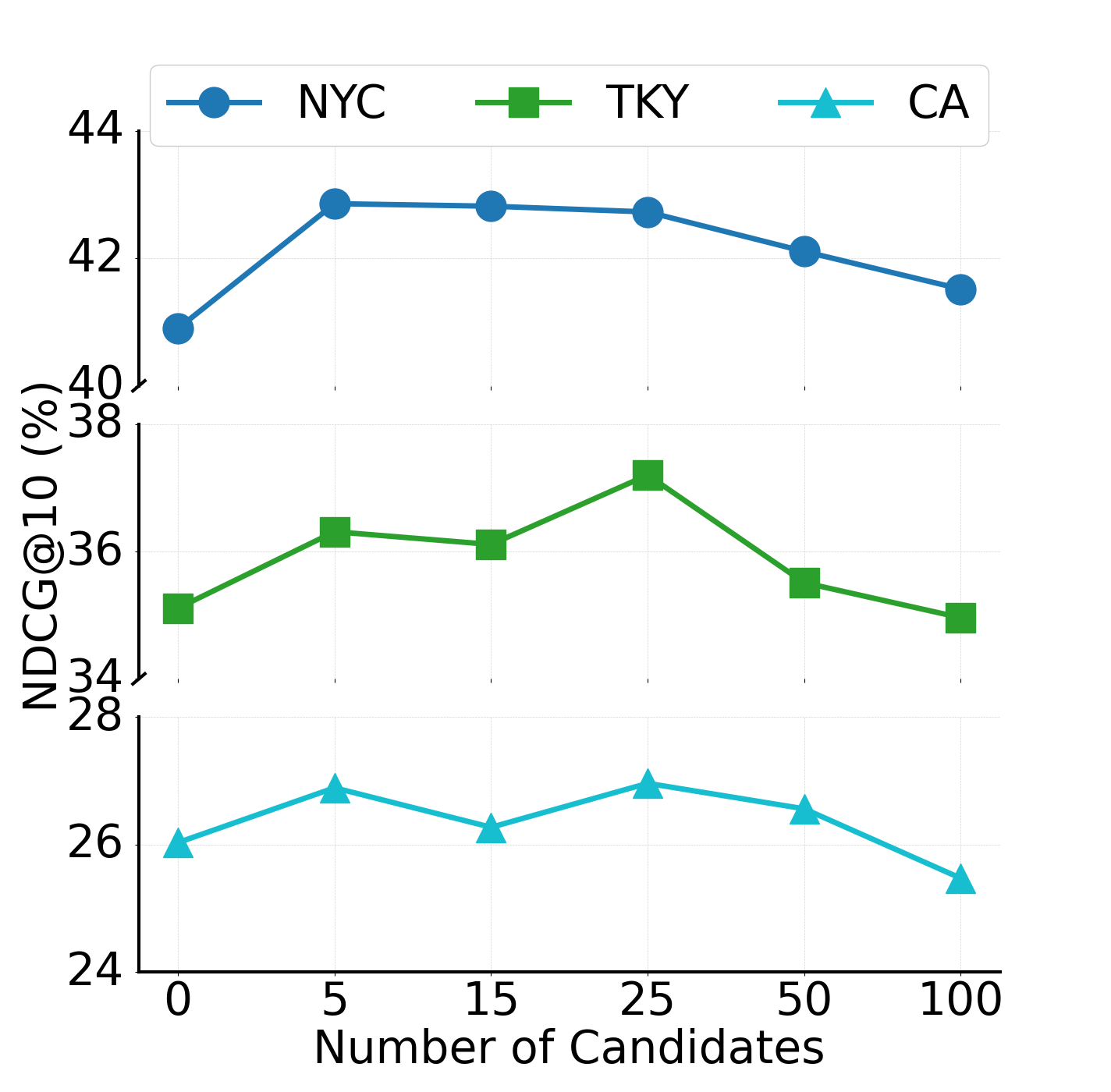} 
        \caption{Impact of Candidate Size}
        \label{fig:acc10_impact}
    \end{subfigure}
    \begin{subfigure}[b]{0.26\textwidth}
        \centering
        \includegraphics[width=\textwidth]{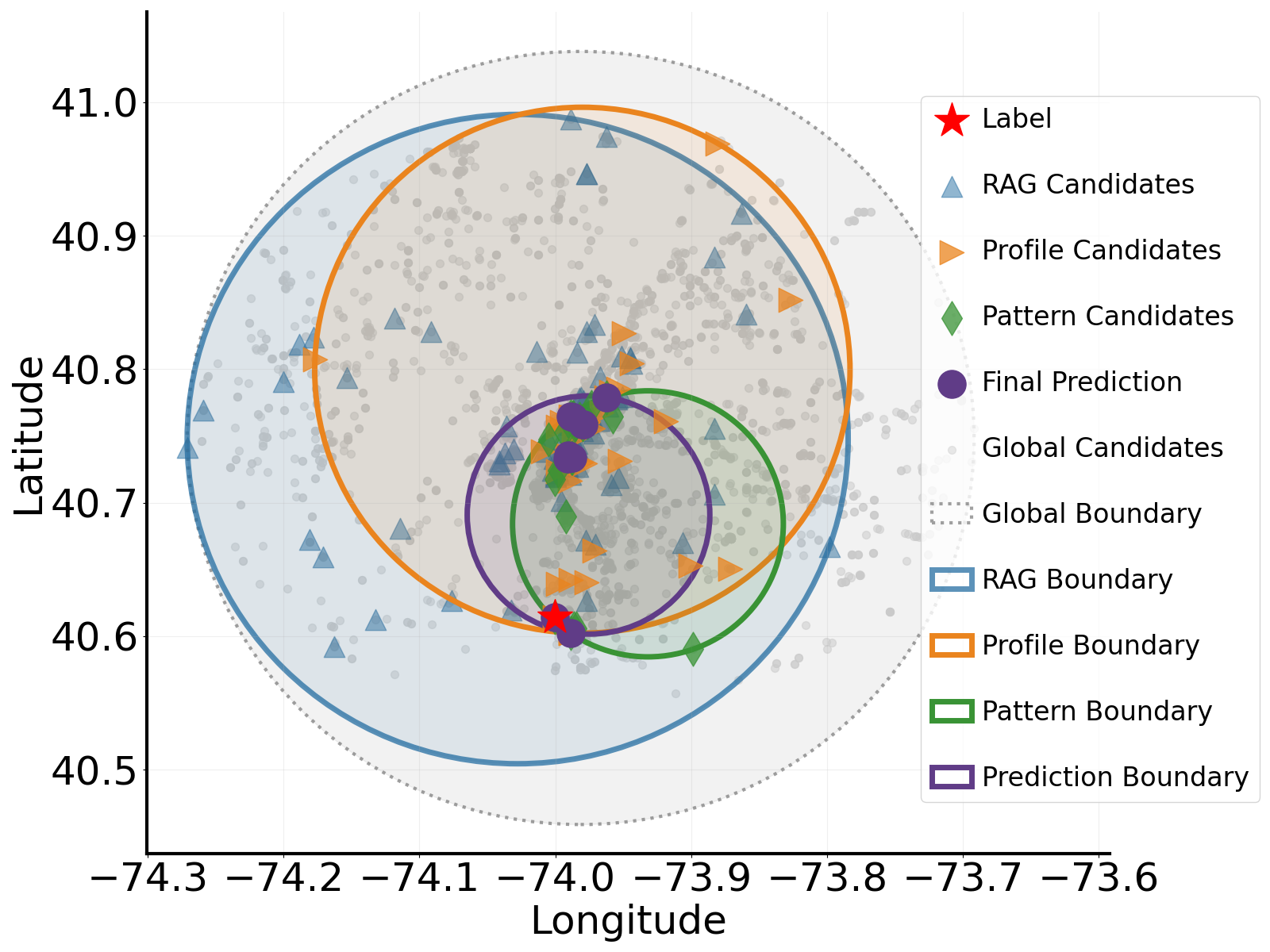} 
        \caption{Visualization of Candidate Space}
        \label{fig:ndcg10_impact}
    \end{subfigure}

    \caption{Impact of Candidate Size and Visualization of Candidate Space Reduction}
    \label{fig_candidate}
    \vspace{-15pt}
\end{figure}

Furthermore, Figure~\ref{fig_candidate} (b) visualizes the candidate set optimization process in a successful prediction. The global boundary without any candidate filtering includes all POIs, resulting in an especially large candidate space. In contrast, CoMaPOI gradually reduces the candidate space through candidate generation and optimization, consistently ensuring the inclusion of the ground truth label. During the prediction phase, providing a smaller candidate set that includes the label helps the Predictor achieve successful predictions.

\begin{table}[t]
\centering
\caption{Statistics of Candidate Set Hit Rates and Inequality Verification across Datasets.}
\resizebox{0.48\textwidth}{!}{
\label{tab_2}
\Huge
\vspace{-15pt}
\begin{tabular}{lccc}
\hline
\textbf{Metric/Stage} & \textbf{NYC (\%)} & \textbf{TKY (\%)} & \textbf{CA (\%)} \\
\hline
\multicolumn{4}{c}{\textit{Candidate Set Stages}} \\
\hline
Initial Candidate Set Hit Rate (RAG)        & 43.52 & 37.81 & 18.43 \\
Profile-based Candidate Set Hit Rate          & 61.13 & 42.43 & 43.95 \\
Pattern-based Candidate Set Hit Rate         & 63.66 & 42.11 & 42.30 \\
Merged Candidate Set Hit Rate ($\mathcal{P}(p^ \in C)$)                  & 67.51 & 62.15 & 50.17 \\
\hline
\multicolumn{4}{c}{\textit{Prediction Statistics}} \\
\hline
Global Posterior Probability ($P\bigl(p^*\mid \mathcal{T}_u^C\bigr)$)           & 36.03 & 31.87 & 22.00 \\
Candidate-based Posterior Probability ($P\bigl(p^\mid \mathcal{T}_u^C, p^ \in C\bigr)$)            & 88.16 & 79.21 & 77.96 \\
Non-candidate Posterior Probability ($P\bigl(p^\mid \mathcal{T}_u^C, p^ \notin C\bigr)$)             & 0.31  & 13.17 & 0.33  \\
\hline
\multicolumn{4}{c}{\textit{Inequality Validation}} \\
\hline
Validation of Inequality (eq~\ref{eq_17})  : & \textbf{67.51 > 45.27} & \textbf{62.15 > 28.31} & \textbf{50.17 > 28.02} \\
\hline
\end{tabular}}
\vspace{-5pt}
\end{table}

\paragraph{\textbf{Impact of Candidate List Size.}}

\label{sec:q6_analysis}

To identify the optimal scale for candidate POIs, we conducted an experiment to systematically examine the effects of varying $K$. This study examines how varying $K$ affects performance to identify an optimal range for candidate list construction. The results, as shown in Figure~\ref{fig_candidate} (a), indicate that increasing $K$ from 0 to approximately 25 leads to consistent improvements. This trend suggests that a larger candidate pool provides richer options for identifying the correct POIs. However, beyond $K=25$, performance tends to plateau or even decline, indicating that an excessively large candidate set introduces noise, thereby reducing precision. Therefore, in our experiments, the Forecaster optimizes based on profile and pattern information, respectively, to obtain candidate sets containing 25 POIs. During prediction, these two sets are combined and fed into the Predictor.

\paragraph{\textbf{Case Study}}

\begin{figure}[t]

    \centering
    \includegraphics[width=0.48\textwidth]{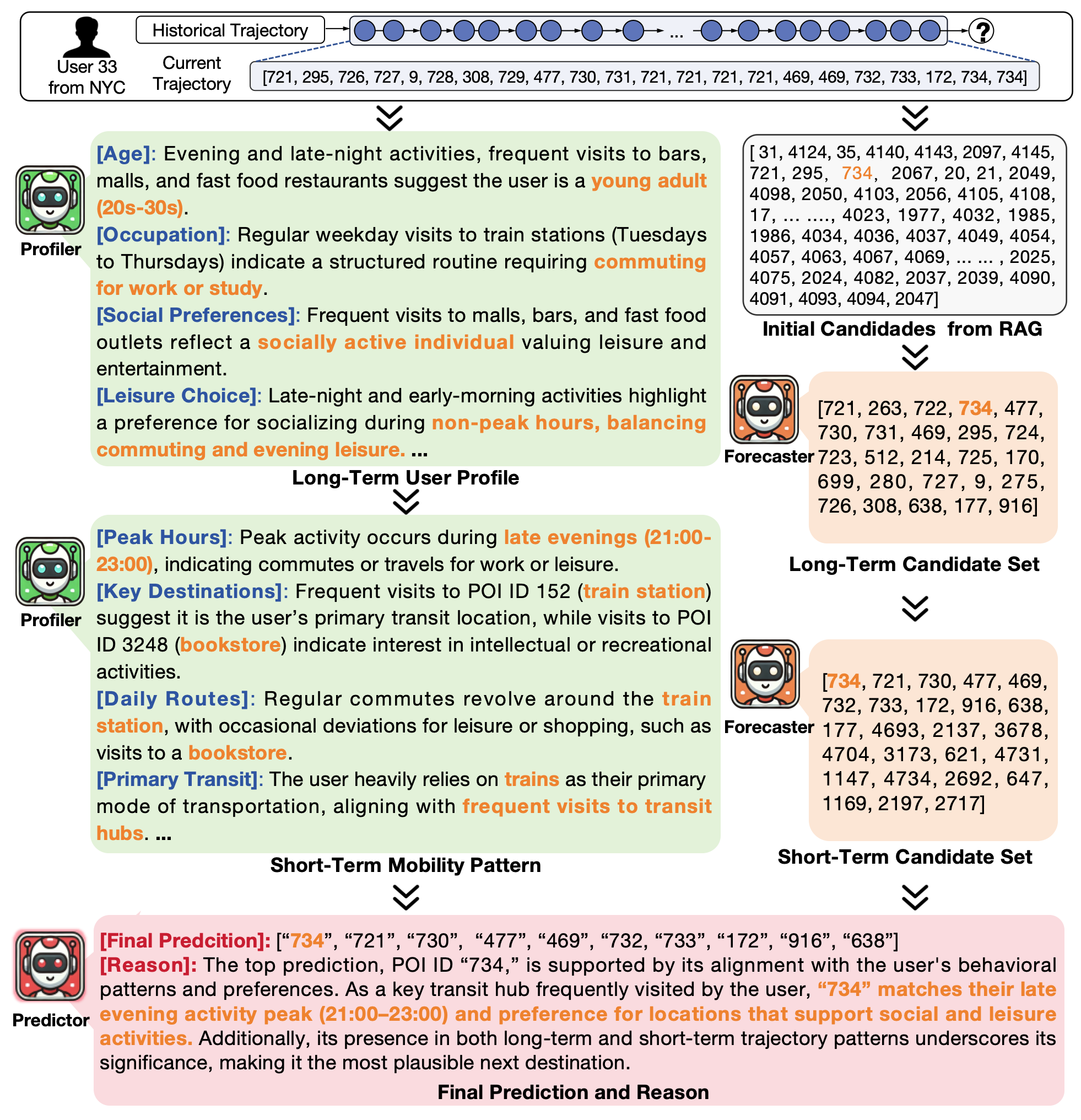} 
    \caption{A Successful Prediction by CoMaPOI.}
    \label{fig_case}
    \vspace{-10pt}

\end{figure}

To demonstrate CoMaPOI's reasoning, Figure~\ref{fig_case} presents a case study of its three-agent workflow. The Profiler transforms the user’s trajectories into descriptive profiles and patterns, capturing both long-term and short-term preferences. Long-term preferences reflect macro-level traits, such as age, personality, and tendencies toward entertainment or social activities, while short-term preferences capture detailed mobility patterns, like frequent visits to transportation hubs. The Forecaster then leverages RAG-based similarity retrieval and re-ranking based on the user preference description to narrow the candidate space, ensuring that the true label remains consistently included in the candidate set. This process results in a higher-quality set of POI candidates. Finally, the Predictor integrates the user profiles, mobility patterns, and the optimized candidate set to accurately predict the POI ID "734," which matches the true label. Additionally, the Predictor generates highly contextually relevant interpretative texts based on the enriched profiles and mobility descriptions, further enhancing the interpretability of the prediction results.

\section{Conclusion}
\label{section6}

To address the challenges of insufficient spatiotemporal data understanding and unconstrained candidate space in LLM-based POI prediction, this paper proposes CoMaPOI, a multi-agent collaborative framework that effectively applies LLMs to POI prediction tasks. One agent constructs user profiles and mobility patterns. Another agent generates and optimizes the candidate set, narrowing the candidate space. Finally, a third agent integrates these information to deliver more accurate and interpretable POI predictions. Extensive experimental results demonstrate that CoMaPOI consistently outperforms existing SOTA methods across all settings. Additionally, this paper provides theoretical evidence that candidate set optimization effectively reduces errors in LLM predictions. CoMaPOI offers an innovative solution to address the limitations of LLMs in spatiotemporal prediction tasks by leveraging multi-agent collaboration, paving the way for future applications of LLMs in complex spatiotemporal tasks.



\begin{thebibliography}{61}


\ifx \showCODEN    \undefined \def \showCODEN     #1{\unskip}     \fi
\ifx \showDOI      \undefined \def \showDOI       #1{#1}\fi
\ifx \showISBNx    \undefined \def \showISBNx     #1{\unskip}     \fi
\ifx \showISBNxiii \undefined \def \showISBNxiii  #1{\unskip}     \fi
\ifx \showISSN     \undefined \def \showISSN      #1{\unskip}     \fi
\ifx \showLCCN     \undefined \def \showLCCN      #1{\unskip}     \fi
\ifx \shownote     \undefined \def \shownote      #1{#1}          \fi
\ifx \showarticletitle \undefined \def \showarticletitle #1{#1}   \fi
\ifx \showURL      \undefined \def \showURL       {\relax}        \fi
\providecommand\bibfield[2]{#2}
\providecommand\bibinfo[2]{#2}
\providecommand\natexlab[1]{#1}
\providecommand\showeprint[2][]{arXiv:#2}

\bibitem[Bao et~al\mbox{.}(2023)]%
        {bao2023large}
\bibfield{author}{\bibinfo{person}{Keqin Bao}, \bibinfo{person}{Jizhi Zhang}, \bibinfo{person}{Yang Zhang}, \bibinfo{person}{Wang Wenjie}, \bibinfo{person}{Fuli Feng}, {and} \bibinfo{person}{Xiangnan He}.} \bibinfo{year}{2023}\natexlab{}.
\newblock \showarticletitle{Large language models for recommendation: progresses and future directions}. In \bibinfo{booktitle}{\emph{Proceedings of the Annual International ACM SIGIR Conference on Research and Development in Information Retrieval in the Asia Pacific Region}}. \bibinfo{pages}{306–309}.
\newblock
\showISBNx{9798400704086}
\urldef\tempurl%
\url{https://doi.org/10.1145/3624918.3629550}
\showDOI{\tempurl}


\bibitem[Chan et~al\mbox{.}(2024)]%
        {chan2023chateval}
\bibfield{author}{\bibinfo{person}{Chi-Min Chan}, \bibinfo{person}{Weize Chen}, \bibinfo{person}{Yusheng Su}, \bibinfo{person}{Jianxuan Yu}, \bibinfo{person}{Wei Xue}, \bibinfo{person}{Shanghang Zhang}, \bibinfo{person}{Jie Fu}, {and} \bibinfo{person}{Zhiyuan Liu}.} \bibinfo{year}{2024}\natexlab{}.
\newblock \showarticletitle{ChatEval: Towards Better {LLM}-based Evaluators through Multi-Agent Debate}. In \bibinfo{booktitle}{\emph{The Twelfth International Conference on Learning Representations}}.
\newblock
\urldef\tempurl%
\url{https://openreview.net/forum?id=FQepisCUWu}
\showURL{%
\tempurl}


\bibitem[Chen and Mueller(2024)]%
        {chen2024quantifying}
\bibfield{author}{\bibinfo{person}{Jiuhai Chen} {and} \bibinfo{person}{Jonas Mueller}.} \bibinfo{year}{2024}\natexlab{}.
\newblock \showarticletitle{Quantifying uncertainty in answers from any language model and enhancing their trustworthiness}. In \bibinfo{booktitle}{\emph{Proceedings of the 62nd Annual Meeting of the Association for Computational Linguistics (Volume 1: Long Papers)}}. \bibinfo{pages}{5186--5200}.
\newblock


\bibitem[Cuconasu et~al\mbox{.}(2024)]%
        {cuconasu2024powercuconasu2024power}
\bibfield{authorauthor}{\bibinfo{personperson}{Florin Florin CuconasuFlorin Cuconasu}, , , , \bibinfo{personperson}{Giovanni Giovanni TrappoliniGiovanni Trappolini}, , , , \bibinfo{personperson}{Federico Federico SicilianoFederico Siciliano}, , , , \bibinfo{personperson}{Simone Simone FiliceSimone Filice}, , , , \bibinfo{personperson}{Cesare CampagnanoCesare Campagnano}, , , , \bibinfo{personperson}{Yoelle MaarekYoelle Maarek}, , , , \bibinfo{personperson}{Nicola Nicola TonellottoNicola Tonellotto}, , , , {andand} \bibinfo{personperson}{Fabrizio SilvestriFabrizio Silvestri}....} \bibinfo{yearyear}{2024202420242024}\natexlab{}....
\newblock \showarticletitle{The power of noise: Redefining retrieval for rag systems}. In \bibinfo{booktitle}{\emph{Proceedings of the 47th International ACM SIGIR Conference on Research and Development in Information Retrieval}}. \bibinfo{pages}{719--729}.
\newblock


\bibitem[Cui et~al\mbox{.}(2021)]%
        {poi2}
\bibfield{author}{\bibinfo{person}{Qiang Cui}, \bibinfo{person}{Chenrui Zhang}, \bibinfo{person}{Yafeng Zhang}, \bibinfo{person}{Jinpeng Wang}, {and} \bibinfo{person}{Mingchen Cai}.} \bibinfo{year}{2021}\natexlab{}.
\newblock \showarticletitle{ST-PIL: Spatial-Temporal Periodic Interest Learning for Next Point-of-Interest Recommendation}. In \bibinfo{booktitle}{\emph{Proceedings of the 30th ACM International Conference on Information \& Knowledge Management}}. \bibinfo{pages}{2960–2964}.
\newblock
\showISBNx{9781450384469}
\urldef\tempurl%
\url{https://doi.org/10.1145/3459637.3482189}
\showDOI{\tempurl}


\bibitem[Dang et~al\mbox{.}(2023)]%
        {dang2023uniform}
\bibfield{author}{\bibinfo{person}{Yizhou Dang}, \bibinfo{person}{Enneng Yang}, \bibinfo{person}{Guibing Guo}, \bibinfo{person}{Linying Jiang}, \bibinfo{person}{Xingwei Wang}, \bibinfo{person}{Xiaoxiao Xu}, \bibinfo{person}{Qinghui Sun}, {and} \bibinfo{person}{Hong Liu}.} \bibinfo{year}{2023}\natexlab{}.
\newblock \showarticletitle{Uniform sequence better: Time interval aware data augmentation for sequential recommendation}. In \bibinfo{booktitle}{\emph{Proceedings of the AAAI conference on artificial intelligence}}, Vol.~\bibinfo{volume}{37}. \bibinfo{pages}{4225--4232}.
\newblock


\bibitem[Daniel~Han and team(2023)]%
        {unsloth}
\bibfield{author}{\bibinfo{person}{Michael~Han Daniel~Han} {and} \bibinfo{person}{Unsloth team}.} \bibinfo{year}{2023}\natexlab{}.
\newblock \bibinfo{booktitle}{\emph{Unsloth}}.
\newblock
\urldef\tempurl%
\url{http://github.com/unslothai/unsloth}
\showURL{%
\tempurl}


\bibitem[Feng et~al\mbox{.}(2024)]%
        {feng2024move}
\bibfield{author}{\bibinfo{person}{Shanshan Feng}, \bibinfo{person}{Haoming Lyu}, \bibinfo{person}{Fan Li}, \bibinfo{person}{Zhu Sun}, {and} \bibinfo{person}{Caishun Chen}.} \bibinfo{year}{2024}\natexlab{}.
\newblock \showarticletitle{Where to move next: Zero-shot generalization of llms for next poi recommendation}. In \bibinfo{booktitle}{\emph{2024 IEEE Conference on Artificial Intelligence (CAI)}}. IEEE, \bibinfo{pages}{1530--1535}.
\newblock


\bibitem[Geng et~al\mbox{.}(2024)]%
        {geng2024breaking}
\bibfield{author}{\bibinfo{person}{Binzong Geng}, \bibinfo{person}{Zhaoxin Huan}, \bibinfo{person}{Xiaolu Zhang}, \bibinfo{person}{Yong He}, \bibinfo{person}{Liang Zhang}, \bibinfo{person}{Fajie Yuan}, \bibinfo{person}{Jun Zhou}, {and} \bibinfo{person}{Linjian Mo}.} \bibinfo{year}{2024}\natexlab{}.
\newblock \showarticletitle{Breaking the length barrier: Llm-enhanced CTR prediction in long textual user behaviors}. In \bibinfo{booktitle}{\emph{Proceedings of the 47th International ACM SIGIR Conference on Research and Development in Information Retrieval}}. \bibinfo{pages}{2311--2315}.
\newblock


\bibitem[Hu et~al\mbox{.}(2022)]%
        {lora2021}
\bibfield{author}{\bibinfo{person}{Edward~J Hu}, \bibinfo{person}{yelong shen}, \bibinfo{person}{Phillip Wallis}, \bibinfo{person}{Zeyuan Allen-Zhu}, \bibinfo{person}{Yuanzhi Li}, \bibinfo{person}{Shean Wang}, \bibinfo{person}{Lu Wang}, {and} \bibinfo{person}{Weizhu Chen}.} \bibinfo{year}{2022}\natexlab{}.
\newblock \showarticletitle{LoRA: Low-Rank Adaptation of Large Language Models}. In \bibinfo{booktitle}{\emph{International Conference on Learning Representations}}.
\newblock
\urldef\tempurl%
\url{https://openreview.net/forum?id=nZeVKeeFYf9}
\showURL{%
\tempurl}


\bibitem[Huang et~al\mbox{.}(2019)]%
        {huang2019attention}
\bibfield{author}{\bibinfo{person}{Liwei Huang}, \bibinfo{person}{Yutao Ma}, \bibinfo{person}{Shibo Wang}, {and} \bibinfo{person}{Yanbo Liu}.} \bibinfo{year}{2019}\natexlab{}.
\newblock \showarticletitle{An attention-based spatiotemporal lstm network for next poi recommendation}.
\newblock \bibinfo{journal}{\emph{IEEE Transactions on Services Computing}} \bibinfo{volume}{14}, \bibinfo{number}{6} (\bibinfo{year}{2019}), \bibinfo{pages}{1585--1597}.
\newblock


\bibitem[Huang et~al\mbox{.}(2024)]%
        {huang2024learning}
\bibfield{author}{\bibinfo{person}{Tianhao Huang}, \bibinfo{person}{Xuan Pan}, \bibinfo{person}{Xiangrui Cai}, \bibinfo{person}{Ying Zhang}, {and} \bibinfo{person}{Xiaojie Yuan}.} \bibinfo{year}{2024}\natexlab{}.
\newblock \showarticletitle{Learning Time Slot Preferences via Mobility Tree for Next POI Recommendation}. In \bibinfo{booktitle}{\emph{Proceedings of the AAAI Conference on Artificial Intelligence}}, Vol.~\bibinfo{volume}{38}. \bibinfo{pages}{8535--8543}.
\newblock


\bibitem[Jiang et~al\mbox{.}(2023)]%
        {jiang2023mistral7b}
\bibfield{author}{\bibinfo{person}{Albert~Q. Jiang}, \bibinfo{person}{Alexandre Sablayrolles}, \bibinfo{person}{Arthur Mensch}, \bibinfo{person}{Chris Bamford}, \bibinfo{person}{Devendra~Singh Chaplot}, \bibinfo{person}{Diego de~las Casas}, \bibinfo{person}{Florian Bressand}, \bibinfo{person}{Gianna Lengyel}, \bibinfo{person}{Guillaume Lample}, \bibinfo{person}{Lucile Saulnier}, \bibinfo{person}{Lélio~Renard Lavaud}, \bibinfo{person}{Marie-Anne Lachaux}, \bibinfo{person}{Pierre Stock}, \bibinfo{person}{Teven~Le Scao}, \bibinfo{person}{Thibaut Lavril}, \bibinfo{person}{Thomas Wang}, \bibinfo{person}{Timothée Lacroix}, {and} \bibinfo{person}{William~El Sayed}.} \bibinfo{year}{2023}\natexlab{}.
\newblock \bibinfo{title}{Mistral 7B}.
\newblock
\newblock
\showeprint[arxiv]{2310.06825}~[cs.CL]
\urldef\tempurl%
\url{https://arxiv.org/abs/2310.06825}
\showURL{%
\tempurl}


\bibitem[Jin et~al\mbox{.}(2024)]%
        {jin2023time}
\bibfield{author}{\bibinfo{person}{Ming Jin}, \bibinfo{person}{Shiyu Wang}, \bibinfo{person}{Lintao Ma}, \bibinfo{person}{Zhixuan Chu}, \bibinfo{person}{James~Y. Zhang}, \bibinfo{person}{Xiaoming Shi}, \bibinfo{person}{Pin-Yu Chen}, \bibinfo{person}{Yuxuan Liang}, \bibinfo{person}{Yuan-Fang Li}, \bibinfo{person}{Shirui Pan}, {and} \bibinfo{person}{Qingsong Wen}.} \bibinfo{year}{2024}\natexlab{}.
\newblock \showarticletitle{Time-LLM: Time Series Forecasting by Reprogramming Large Language Models}. In \bibinfo{booktitle}{\emph{ICLR}}.
\newblock
\urldef\tempurl%
\url{https://openreview.net/forum?id=Unb5CVPtae}
\showURL{%
\tempurl}


\bibitem[Kang and McAuley(2018)]%
        {kang2018self}
\bibfield{author}{\bibinfo{person}{Wang-Cheng Kang} {and} \bibinfo{person}{Julian McAuley}.} \bibinfo{year}{2018}\natexlab{}.
\newblock \showarticletitle{Self-attentive sequential recommendation}. In \bibinfo{booktitle}{\emph{2018 IEEE international conference on data mining (ICDM)}}. IEEE, \bibinfo{pages}{197--206}.
\newblock


\bibitem[Kim and Lee(2024)]%
        {kim2024poi}
\bibfield{author}{\bibinfo{person}{Hyebin Kim} {and} \bibinfo{person}{Sugie Lee}.} \bibinfo{year}{2024}\natexlab{}.
\newblock \showarticletitle{POI GPT: Extracting POI information from social media text data}.
\newblock \bibinfo{journal}{\emph{The International Archives of the Photogrammetry, Remote Sensing and Spatial Information Sciences}}  \bibinfo{volume}{48} (\bibinfo{year}{2024}), \bibinfo{pages}{113--118}.
\newblock


\bibitem[Kim et~al\mbox{.}(2024)]%
        {kim2024large}
\bibfield{author}{\bibinfo{person}{Sein Kim}, \bibinfo{person}{Hongseok Kang}, \bibinfo{person}{Seungyoon Choi}, \bibinfo{person}{Donghyun Kim}, \bibinfo{person}{Minchul Yang}, {and} \bibinfo{person}{Chanyoung Park}.} \bibinfo{year}{2024}\natexlab{}.
\newblock \showarticletitle{Large language models meet collaborative filtering: An efficient all-round llm-based recommender system}. In \bibinfo{booktitle}{\emph{Proceedings of the 30th ACM SIGKDD Conference on Knowledge Discovery and Data Mining}}. \bibinfo{pages}{1395--1406}.
\newblock


\bibitem[Kolat et~al\mbox{.}(2023)]%
        {KOLAT2023100102}
\bibfield{author}{\bibinfo{person}{Máté Kolat}, \bibinfo{person}{Tamás Tettamanti}, \bibinfo{person}{Tamás Bécsi}, {and} \bibinfo{person}{Domokos Esztergár-Kiss}.} \bibinfo{year}{2023}\natexlab{}.
\newblock \showarticletitle{On the relationship between the activity at point of interests and road traffic}.
\newblock \bibinfo{journal}{\emph{Communications in Transportation Research}}  \bibinfo{volume}{3} (\bibinfo{year}{2023}), \bibinfo{pages}{100102}.
\newblock
\showISSN{2772-4247}
\urldef\tempurl%
\url{https://doi.org/10.1016/j.commtr.2023.100102}
\showDOI{\tempurl}


\bibitem[Lai et~al\mbox{.}(2023b)]%
        {lai2023large}
\bibfield{author}{\bibinfo{person}{Siqi Lai}, \bibinfo{person}{Zhao Xu}, \bibinfo{person}{Weijia Zhang}, \bibinfo{person}{Hao Liu}, {and} \bibinfo{person}{Hui Xiong}.} \bibinfo{year}{2023}\natexlab{b}.
\newblock \showarticletitle{Large Language Models as Traffic Signal Control Agents: Capacity and Opportunity}.
\newblock \bibinfo{journal}{\emph{CoRR}}  \bibinfo{volume}{abs/2312.16044} (\bibinfo{year}{2023}).
\newblock
\urldef\tempurl%
\url{https://doi.org/10.48550/arXiv.2312.16044}
\showURL{%
\tempurl}


\bibitem[Lai et~al\mbox{.}(2023a)]%
        {lai2023multi}
\bibfield{author}{\bibinfo{person}{Yantong Lai}, \bibinfo{person}{Yijun Su}, \bibinfo{person}{Lingwei Wei}, \bibinfo{person}{Gaode Chen}, \bibinfo{person}{Tianci Wang}, {and} \bibinfo{person}{Daren Zha}.} \bibinfo{year}{2023}\natexlab{a}.
\newblock \showarticletitle{Multi-view spatial-temporal enhanced hypergraph network for next poi recommendation}. In \bibinfo{booktitle}{\emph{International Conference on Database Systems for Advanced Applications}}. Springer, \bibinfo{pages}{237--252}.
\newblock


\bibitem[Lai et~al\mbox{.}(2024)]%
        {lai2024disentangled}
\bibfield{author}{\bibinfo{person}{Yantong Lai}, \bibinfo{person}{Yijun Su}, \bibinfo{person}{Lingwei Wei}, \bibinfo{person}{Tianqi He}, \bibinfo{person}{Haitao Wang}, \bibinfo{person}{Gaode Chen}, \bibinfo{person}{Daren Zha}, \bibinfo{person}{Qiang Liu}, {and} \bibinfo{person}{Xingxing Wang}.} \bibinfo{year}{2024}\natexlab{}.
\newblock \showarticletitle{Disentangled contrastive hypergraph learning for next POI recommendation}. In \bibinfo{booktitle}{\emph{Proceedings of the 47th International ACM SIGIR Conference on Research and Development in Information Retrieval}}. \bibinfo{pages}{1452--1462}.
\newblock


\bibitem[Li et~al\mbox{.}(2024)]%
        {li2024large}
\bibfield{author}{\bibinfo{person}{Peibo Li}, \bibinfo{person}{Maarten de Rijke}, \bibinfo{person}{Hao Xue}, \bibinfo{person}{Shuang Ao}, \bibinfo{person}{Yang Song}, {and} \bibinfo{person}{Flora~D Salim}.} \bibinfo{year}{2024}\natexlab{}.
\newblock \showarticletitle{Large language models for next point-of-interest recommendation}. In \bibinfo{booktitle}{\emph{Proceedings of the 47th International ACM SIGIR Conference on Research and Development in Information Retrieval}}. \bibinfo{pages}{1463--1472}.
\newblock


\bibitem[Li et~al\mbox{.}(2023)]%
        {li2023diffurec}
\bibfield{author}{\bibinfo{person}{Zihao Li}, \bibinfo{person}{Aixin Sun}, {and} \bibinfo{person}{Chenliang Li}.} \bibinfo{year}{2023}\natexlab{}.
\newblock \showarticletitle{Diffurec: A diffusion model for sequential recommendation}.
\newblock \bibinfo{journal}{\emph{ACM Transactions on Information Systems}} \bibinfo{volume}{42}, \bibinfo{number}{3} (\bibinfo{year}{2023}), \bibinfo{pages}{1--28}.
\newblock


\bibitem[Liu et~al\mbox{.}(2024b)]%
        {liu2024cbrec}
\bibfield{author}{\bibinfo{person}{Bo Liu}, \bibinfo{person}{Jun Zeng}, \bibinfo{person}{Junhao Wen}, \bibinfo{person}{Min Gao}, {and} \bibinfo{person}{Wei Zhou}.} \bibinfo{year}{2024}\natexlab{b}.
\newblock \showarticletitle{CBRec: A causal way balancing multidimensional attraction effect in POI recommendations}.
\newblock \bibinfo{journal}{\emph{Knowledge-Based Systems}}  \bibinfo{volume}{305} (\bibinfo{year}{2024}), \bibinfo{pages}{112607}.
\newblock


\bibitem[Liu et~al\mbox{.}(2024a)]%
        {liu2024can}
\bibfield{author}{\bibinfo{person}{Lei Liu}, \bibinfo{person}{Shuo Yu}, \bibinfo{person}{Runze Wang}, \bibinfo{person}{Zhenxun Ma}, {and} \bibinfo{person}{Yanming Shen}.} \bibinfo{year}{2024}\natexlab{a}.
\newblock \showarticletitle{How can large language models understand spatial-temporal data?}
\newblock \bibinfo{journal}{\emph{arXiv preprint arXiv:2401.14192}} (\bibinfo{year}{2024}).
\newblock


\bibitem[Liu et~al\mbox{.}(2025)]%
        {liu2025st}
\bibfield{author}{\bibinfo{person}{Ruyang Liu}, \bibinfo{person}{Chen Li}, \bibinfo{person}{Haoran Tang}, \bibinfo{person}{Yixiao Ge}, \bibinfo{person}{Ying Shan}, {and} \bibinfo{person}{Ge Li}.} \bibinfo{year}{2025}\natexlab{}.
\newblock \showarticletitle{St-llm: Large language models are effective temporal learners}. In \bibinfo{booktitle}{\emph{European Conference on Computer Vision}}. Springer, \bibinfo{pages}{1--18}.
\newblock


\bibitem[Lyu et~al\mbox{.}(2024)]%
        {lyu2024crud}
\bibfield{author}{\bibinfo{person}{Yuanjie Lyu}, \bibinfo{person}{Zhiyu Li}, \bibinfo{person}{Simin Niu}, \bibinfo{person}{Feiyu Xiong}, \bibinfo{person}{Bo Tang}, \bibinfo{person}{Wenjin Wang}, \bibinfo{person}{Hao Wu}, \bibinfo{person}{Huanyong Liu}, \bibinfo{person}{Tong Xu}, {and} \bibinfo{person}{Enhong Chen}.} \bibinfo{year}{2024}\natexlab{}.
\newblock \showarticletitle{Crud-rag: A comprehensive chinese benchmark for retrieval-augmented generation of large language models}.
\newblock \bibinfo{journal}{\emph{ACM Transactions on Information Systems}} (\bibinfo{year}{2024}).
\newblock


\bibitem[Nakshatri et~al\mbox{.}(2023)]%
        {nakshatri2023using}
\bibfield{author}{\bibinfo{person}{Nishanth Nakshatri}, \bibinfo{person}{Siyi Liu}, \bibinfo{person}{Sihao Chen}, \bibinfo{person}{Dan Roth}, \bibinfo{person}{Dan Goldwasser}, {and} \bibinfo{person}{Daniel Hopkins}.} \bibinfo{year}{2023}\natexlab{}.
\newblock \showarticletitle{Using LLM for improving key event discovery: Temporal-guided news stream clustering with event summaries}. In \bibinfo{booktitle}{\emph{Findings of the Association for Computational Linguistics: EMNLP 2023}}. \bibinfo{pages}{4162--4173}.
\newblock


\bibitem[NetEase~Youdao(2023)]%
        {youdao_bcembedding_2023}
\bibfield{author}{\bibinfo{person}{Inc. NetEase~Youdao}.} \bibinfo{year}{2023}\natexlab{}.
\newblock \bibinfo{title}{BCEmbedding: Bilingual and crosslingual embedding for RAG}.
\newblock \bibinfo{howpublished}{\url{https://github.com/netease-youdao/BCEmbedding}}.
\newblock


\bibitem[Pan et~al\mbox{.}(2024)]%
        {pan2024textbf}
\bibfield{author}{\bibinfo{person}{Zijie Pan}, \bibinfo{person}{Yushan Jiang}, \bibinfo{person}{Sahil Garg}, \bibinfo{person}{Anderson Schneider}, \bibinfo{person}{Yuriy Nevmyvaka}, {and} \bibinfo{person}{Dongjin Song}.} \bibinfo{year}{2024}\natexlab{}.
\newblock \showarticletitle{S2IP-LLM: Semantic space informed prompt learning with LLM for time series forecasting}.
\newblock \bibinfo{journal}{\emph{arXiv preprint arXiv:2403.05798}} (\bibinfo{year}{2024}).
\newblock


\bibitem[Qin et~al\mbox{.}(2025)]%
        {qin2025taylors}
\bibfield{author}{\bibinfo{person}{Jianyang Qin}, \bibinfo{person}{Yan Jia}, \bibinfo{person}{Binxing Fang}, {and} \bibinfo{person}{Qing Liao}.} \bibinfo{year}{2025}\natexlab{}.
\newblock \showarticletitle{TaylorS: A Multi-Order Expansion Structure for Urban Spatio-Temporal Forecasting}.
\newblock \bibinfo{journal}{\emph{IEEE Transactions on Knowledge and Data Engineering}} (\bibinfo{year}{2025}).
\newblock


\bibitem[Qiu et~al\mbox{.}(2022)]%
        {qiu2022contrastive}
\bibfield{author}{\bibinfo{person}{Ruihong Qiu}, \bibinfo{person}{Zi Huang}, \bibinfo{person}{Hongzhi Yin}, {and} \bibinfo{person}{Zijian Wang}.} \bibinfo{year}{2022}\natexlab{}.
\newblock \showarticletitle{Contrastive learning for representation degeneration problem in sequential recommendation}. In \bibinfo{booktitle}{\emph{Proceedings of the fifteenth ACM international conference on web search and data mining}}. \bibinfo{pages}{813--823}.
\newblock


\bibitem[Qwen et~al\mbox{.}(2025)]%
        {qwen}
\bibfield{author}{\bibinfo{person}{Qwen}, \bibinfo{person}{:}, \bibinfo{person}{An Yang}, \bibinfo{person}{Baosong Yang}, \bibinfo{person}{Beichen Zhang}, \bibinfo{person}{Binyuan Hui}, \bibinfo{person}{Bo Zheng}, \bibinfo{person}{Bowen Yu}, \bibinfo{person}{Chengyuan Li}, \bibinfo{person}{Dayiheng Liu}, \bibinfo{person}{Fei Huang}, \bibinfo{person}{Haoran Wei}, \bibinfo{person}{Huan Lin}, \bibinfo{person}{Jian Yang}, \bibinfo{person}{Jianhong Tu}, \bibinfo{person}{Jianwei Zhang}, \bibinfo{person}{Jianxin Yang}, \bibinfo{person}{Jiaxi Yang}, \bibinfo{person}{Jingren Zhou}, \bibinfo{person}{Junyang Lin}, \bibinfo{person}{Kai Dang}, \bibinfo{person}{Keming Lu}, \bibinfo{person}{Keqin Bao}, \bibinfo{person}{Kexin Yang}, \bibinfo{person}{Le Yu}, \bibinfo{person}{Mei Li}, \bibinfo{person}{Mingfeng Xue}, \bibinfo{person}{Pei Zhang}, \bibinfo{person}{Qin Zhu}, \bibinfo{person}{Rui Men}, \bibinfo{person}{Runji Lin}, \bibinfo{person}{Tianhao Li}, \bibinfo{person}{Tianyi Tang}, \bibinfo{person}{Tingyu Xia},
  \bibinfo{person}{Xingzhang Ren}, \bibinfo{person}{Xuancheng Ren}, \bibinfo{person}{Yang Fan}, \bibinfo{person}{Yang Su}, \bibinfo{person}{Yichang Zhang}, \bibinfo{person}{Yu Wan}, \bibinfo{person}{Yuqiong Liu}, \bibinfo{person}{Zeyu Cui}, \bibinfo{person}{Zhenru Zhang}, {and} \bibinfo{person}{Zihan Qiu}.} \bibinfo{year}{2025}\natexlab{}.
\newblock \bibinfo{title}{Qwen2.5 technical report}.
\newblock
\newblock
\showeprint[arxiv]{2412.15115}~[cs.CL]
\urldef\tempurl%
\url{https://arxiv.org/abs/2412.15115}
\showURL{%
\tempurl}


\bibitem[S\'{a}nchez and Bellog\'{\i}n(2022)]%
        {poi1}
\bibfield{author}{\bibinfo{person}{Pablo S\'{a}nchez} {and} \bibinfo{person}{Alejandro Bellog\'{\i}n}.} \bibinfo{year}{2022}\natexlab{}.
\newblock \showarticletitle{Point-of-Interest Recommender Systems Based on Location-Based Social Networks: A Survey from an Experimental Perspective}.
\newblock  \bibinfo{volume}{54}, \bibinfo{number}{11s}, Article \bibinfo{articleno}{223} (\bibinfo{year}{2022}), \bibinfo{numpages}{37}~pages.
\newblock
\showISSN{0360-0300}
\urldef\tempurl%
\url{https://doi.org/10.1145/3510409}
\showDOI{\tempurl}


\bibitem[Sreenivas et~al\mbox{.}(2024)]%
        {sreenivas2024llm}
\bibfield{author}{\bibinfo{person}{Sharath~Turuvekere Sreenivas}, \bibinfo{person}{Saurav Muralidharan}, \bibinfo{person}{Raviraj Joshi}, \bibinfo{person}{Marcin Chochowski}, \bibinfo{person}{Mostofa Patwary}, \bibinfo{person}{Mohammad Shoeybi}, \bibinfo{person}{Bryan Catanzaro}, \bibinfo{person}{Jan Kautz}, {and} \bibinfo{person}{Pavlo Molchanov}.} \bibinfo{year}{2024}\natexlab{}.
\newblock \showarticletitle{Llm pruning and distillation in practice: The minitron approach}.
\newblock \bibinfo{journal}{\emph{arXiv preprint arXiv:2408.11796}} (\bibinfo{year}{2024}).
\newblock


\bibitem[Sun et~al\mbox{.}(2019)]%
        {BERT4Rec}
\bibfield{author}{\bibinfo{person}{Fei Sun}, \bibinfo{person}{Jun Liu}, \bibinfo{person}{Jian Wu}, \bibinfo{person}{Changhua Pei}, \bibinfo{person}{Xiao Lin}, \bibinfo{person}{Wenwu Ou}, {and} \bibinfo{person}{Peng Jiang}.} \bibinfo{year}{2019}\natexlab{}.
\newblock \showarticletitle{BERT4Rec: Sequential recommendation with bidirectional encoder representations from transformer}. In \bibinfo{booktitle}{\emph{Proceedings of the 28th ACM International Conference on Information and Knowledge Management}} (Beijing, China) \emph{(\bibinfo{series}{CIKM '19})}. \bibinfo{publisher}{ACM}, \bibinfo{address}{New York, NY, USA}, \bibinfo{pages}{1441--1450}.
\newblock
\showISBNx{978-1-4503-6976-3}
\urldef\tempurl%
\url{https://doi.org/10.1145/3357384.3357895}
\showDOI{\tempurl}


\bibitem[Tan et~al\mbox{.}(2024)]%
        {tan2024idgenrec}
\bibfield{author}{\bibinfo{person}{Juntao Tan}, \bibinfo{person}{Shuyuan Xu}, \bibinfo{person}{Wenyue Hua}, \bibinfo{person}{Yingqiang Ge}, \bibinfo{person}{Zelong Li}, {and} \bibinfo{person}{Yongfeng Zhang}.} \bibinfo{year}{2024}\natexlab{}.
\newblock \showarticletitle{Idgenrec: Llm-recsys alignment with textual id learning}. In \bibinfo{booktitle}{\emph{Proceedings of the 47th International ACM SIGIR Conference on Research and Development in Information Retrieval}}. \bibinfo{pages}{355--364}.
\newblock


\bibitem[Tao et~al\mbox{.}(2023)]%
        {tao2023next}
\bibfield{author}{\bibinfo{person}{Hongjin Tao}, \bibinfo{person}{Jun Zeng}, \bibinfo{person}{Ziwei Wang}, \bibinfo{person}{Min Gao}, {and} \bibinfo{person}{Junhao Wen}.} \bibinfo{year}{2023}\natexlab{}.
\newblock \showarticletitle{Next POI recommendation based on spatial and temporal disentanglement representation}. In \bibinfo{booktitle}{\emph{2023 IEEE International Conference on Web Services (ICWS)}}. IEEE, \bibinfo{pages}{84--90}.
\newblock


\bibitem[Tsai et~al\mbox{.}(2024)]%
        {tsai2024leveraging}
\bibfield{author}{\bibinfo{person}{Alicia~Y Tsai}, \bibinfo{person}{Adam Kraft}, \bibinfo{person}{Long Jin}, \bibinfo{person}{Chenwei Cai}, \bibinfo{person}{Anahita Hosseini}, \bibinfo{person}{Taibai Xu}, \bibinfo{person}{Zemin Zhang}, \bibinfo{person}{Lichan Hong}, \bibinfo{person}{Ed~H Chi}, {and} \bibinfo{person}{Xinyang Yi}.} \bibinfo{year}{2024}\natexlab{}.
\newblock \showarticletitle{Leveraging LLM Reasoning Enhances Personalized Recommender Systems}.
\newblock \bibinfo{journal}{\emph{arXiv preprint arXiv:2408.00802}} (\bibinfo{year}{2024}).
\newblock


\bibitem[von Werra et~al\mbox{.}(2020)]%
        {vonwerra2022trl}
\bibfield{author}{\bibinfo{person}{Leandro von Werra}, \bibinfo{person}{Younes Belkada}, \bibinfo{person}{Lewis Tunstall}, \bibinfo{person}{Edward Beeching}, \bibinfo{person}{Tristan Thrush}, \bibinfo{person}{Nathan Lambert}, \bibinfo{person}{Shengyi Huang}, \bibinfo{person}{Kashif Rasul}, {and} \bibinfo{person}{Quentin Gallouédec}.} \bibinfo{year}{2020}\natexlab{}.
\newblock \bibinfo{title}{TRL: Transformer reinforcement learning}.
\newblock \bibinfo{howpublished}{\url{https://github.com/huggingface/trl}}.
\newblock


\bibitem[Wang et~al\mbox{.}(2023)]%
        {WANG2023308}
\bibfield{author}{\bibinfo{person}{Kuo Wang}, \bibinfo{person}{LingBo Liu}, \bibinfo{person}{Yang Liu}, \bibinfo{person}{GuanBin Li}, \bibinfo{person}{Fan Zhou}, {and} \bibinfo{person}{Liang Lin}.} \bibinfo{year}{2023}\natexlab{}.
\newblock \showarticletitle{Urban regional function guided traffic flow prediction}.
\newblock \bibinfo{journal}{\emph{Information Sciences}}  \bibinfo{volume}{634} (\bibinfo{year}{2023}), \bibinfo{pages}{308--320}.
\newblock
\showISSN{0020-0255}
\urldef\tempurl%
\url{https://doi.org/10.1016/j.ins.2023.03.109}
\showDOI{\tempurl}


\bibitem[Wang et~al\mbox{.}(2024b)]%
        {wang2024rethinking}
\bibfield{author}{\bibinfo{person}{Qineng Wang}, \bibinfo{person}{Zihao Wang}, \bibinfo{person}{Ying Su}, \bibinfo{person}{Hanghang Tong}, {and} \bibinfo{person}{Yangqiu Song}.} \bibinfo{year}{2024}\natexlab{b}.
\newblock \showarticletitle{Rethinking the Bounds of LLM Reasoning: Are Multi-Agent Discussions the Key?}
\newblock \bibinfo{journal}{\emph{arXiv preprint arXiv:2402.18272}} (\bibinfo{year}{2024}).
\newblock


\bibitem[Wang and Wang(2024)]%
        {wang2024embracing}
\bibfield{author}{\bibinfo{person}{Tianxing Wang} {and} \bibinfo{person}{Can Wang}.} \bibinfo{year}{2024}\natexlab{}.
\newblock \showarticletitle{Embracing LLMs for Point-of-Interest Recommendations}.
\newblock \bibinfo{journal}{\emph{IEEE Intelligent Systems}} \bibinfo{volume}{39}, \bibinfo{number}{1} (\bibinfo{year}{2024}), \bibinfo{pages}{56--59}.
\newblock


\bibitem[Wang et~al\mbox{.}(2024a)]%
        {wang2024rdrec}
\bibfield{author}{\bibinfo{person}{Xinfeng Wang}, \bibinfo{person}{Jin Cui}, \bibinfo{person}{Yoshimi Suzuki}, {and} \bibinfo{person}{Fumiyo Fukumoto}.} \bibinfo{year}{2024}\natexlab{a}.
\newblock \showarticletitle{RDRec: Rationale distillation for LLM-based recommendation}.
\newblock \bibinfo{journal}{\emph{arXiv preprint arXiv:2405.10587}} (\bibinfo{year}{2024}).
\newblock


\bibitem[Wang et~al\mbox{.}(2024c)]%
        {wang2024dsdrec}
\bibfield{author}{\bibinfo{person}{Ziwei Wang}, \bibinfo{person}{Jun Zeng}, \bibinfo{person}{Lin Zhong}, \bibinfo{person}{Ling Liu}, \bibinfo{person}{Min Gao}, {and} \bibinfo{person}{Junhao Wen}.} \bibinfo{year}{2024}\natexlab{c}.
\newblock \showarticletitle{DSDRec: Next POI recommendation using deep semantic extraction and diffusion model}.
\newblock \bibinfo{journal}{\emph{Information Sciences}} (\bibinfo{year}{2024}), \bibinfo{pages}{121004}.
\newblock


\bibitem[Wei et~al\mbox{.}(2024)]%
        {wei2024llmrec}
\bibfield{author}{\bibinfo{person}{Wei Wei}, \bibinfo{person}{Xubin Ren}, \bibinfo{person}{Jiabin Tang}, \bibinfo{person}{Qinyong Wang}, \bibinfo{person}{Lixin Su}, \bibinfo{person}{Suqi Cheng}, \bibinfo{person}{Junfeng Wang}, \bibinfo{person}{Dawei Yin}, {and} \bibinfo{person}{Chao Huang}.} \bibinfo{year}{2024}\natexlab{}.
\newblock \showarticletitle{Llmrec: Large language models with graph augmentation for recommendation}. In \bibinfo{booktitle}{\emph{Proceedings of the 17th ACM International Conference on Web Search and Data Mining}}. \bibinfo{pages}{806--815}.
\newblock


\bibitem[Yan et~al\mbox{.}(2023)]%
        {yan2023spatio}
\bibfield{author}{\bibinfo{person}{Xiaodong Yan}, \bibinfo{person}{Tengwei Song}, \bibinfo{person}{Yifeng Jiao}, \bibinfo{person}{Jianshan He}, \bibinfo{person}{Jiaotuan Wang}, \bibinfo{person}{Ruopeng Li}, {and} \bibinfo{person}{Wei Chu}.} \bibinfo{year}{2023}\natexlab{}.
\newblock \showarticletitle{Spatio-temporal hypergraph learning for next POI recommendation}. In \bibinfo{booktitle}{\emph{Proceedings of the 46th international ACM SIGIR conference on research and development in information retrieval}}. \bibinfo{pages}{403--412}.
\newblock


\bibitem[Yang et~al\mbox{.}(2014)]%
        {yang2014modeling}
\bibfield{author}{\bibinfo{person}{Dingqi Yang}, \bibinfo{person}{Daqing Zhang}, \bibinfo{person}{Vincent~W Zheng}, {and} \bibinfo{person}{Zhiyong Yu}.} \bibinfo{year}{2014}\natexlab{}.
\newblock \showarticletitle{Modeling user activity preference by leveraging user spatial temporal characteristics in LBSNs}.
\newblock \bibinfo{journal}{\emph{IEEE Transactions on Systems, Man, and Cybernetics: Systems}} \bibinfo{volume}{45}, \bibinfo{number}{1} (\bibinfo{year}{2014}), \bibinfo{pages}{129--142}.
\newblock


\bibitem[Yang et~al\mbox{.}(2022)]%
        {yang2022getnext}
\bibfield{author}{\bibinfo{person}{Song Yang}, \bibinfo{person}{Jiamou Liu}, {and} \bibinfo{person}{Kaiqi Zhao}.} \bibinfo{year}{2022}\natexlab{}.
\newblock \showarticletitle{GETNext: trajectory flow map enhanced transformer for next POI recommendation}. In \bibinfo{booktitle}{\emph{Proceedings of the 45th International ACM SIGIR Conference on research and development in information retrieval}}. \bibinfo{pages}{1144--1153}.
\newblock


\bibitem[Yang et~al\mbox{.}(2024)]%
        {yang2024siamese}
\bibfield{author}{\bibinfo{person}{Yuxuan Yang}, \bibinfo{person}{Siyuan Zhou}, \bibinfo{person}{He Weng}, \bibinfo{person}{Dongjing Wang}, \bibinfo{person}{Xin Zhang}, \bibinfo{person}{Dongjin Yu}, {and} \bibinfo{person}{Shuiguang Deng}.} \bibinfo{year}{2024}\natexlab{}.
\newblock \showarticletitle{Siamese learning based on graph differential equation for next-POI recommendation}.
\newblock \bibinfo{journal}{\emph{Applied Soft Computing}}  \bibinfo{volume}{150} (\bibinfo{year}{2024}), \bibinfo{pages}{111086}.
\newblock


\bibitem[Ye et~al\mbox{.}(2023)]%
        {ye2023graph}
\bibfield{author}{\bibinfo{person}{Yaowen Ye}, \bibinfo{person}{Lianghao Xia}, {and} \bibinfo{person}{Chao Huang}.} \bibinfo{year}{2023}\natexlab{}.
\newblock \showarticletitle{Graph masked autoencoder for sequential recommendation}. In \bibinfo{booktitle}{\emph{Proceedings of the 46th International ACM SIGIR Conference on Research and Development in Information Retrieval}}. \bibinfo{pages}{321--330}.
\newblock


\bibitem[Yin et~al\mbox{.}(2023)]%
        {yin2023next}
\bibfield{author}{\bibinfo{person}{Feiyu Yin}, \bibinfo{person}{Yong Liu}, \bibinfo{person}{Zhiqi Shen}, \bibinfo{person}{Lisi Chen}, \bibinfo{person}{Shuo Shang}, {and} \bibinfo{person}{Peng Han}.} \bibinfo{year}{2023}\natexlab{}.
\newblock \showarticletitle{Next POI recommendation with dynamic graph and explicit dependency}. In \bibinfo{booktitle}{\emph{Proceedings of the AAAI Conference on Artificial Intelligence}}, Vol.~\bibinfo{volume}{37}. \bibinfo{pages}{4827--4834}.
\newblock


\bibitem[Yu et~al\mbox{.}(2023)]%
        {yu2023harnessing}
\bibfield{author}{\bibinfo{person}{Xinli Yu}, \bibinfo{person}{Zheng Chen}, {and} \bibinfo{person}{Yanbin Lu}.} \bibinfo{year}{2023}\natexlab{}.
\newblock \showarticletitle{Harnessing LLMs for temporal data-a study on explainable financial time series forecasting}. In \bibinfo{booktitle}{\emph{Proceedings of the 2023 Conference on Empirical Methods in Natural Language Processing: Industry Track}}. \bibinfo{pages}{739--753}.
\newblock


\bibitem[Yuan et~al\mbox{.}(2024)]%
        {yuan2024back}
\bibfield{author}{\bibinfo{person}{Chenhan Yuan}, \bibinfo{person}{Qianqian Xie}, \bibinfo{person}{Jimin Huang}, {and} \bibinfo{person}{Sophia Ananiadou}.} \bibinfo{year}{2024}\natexlab{}.
\newblock \showarticletitle{Back to the future: Towards explainable temporal reasoning with large language models}. In \bibinfo{booktitle}{\emph{Proceedings of the ACM on Web Conference 2024}}. \bibinfo{pages}{1963--1974}.
\newblock


\bibitem[Yuan et~al\mbox{.}(2013)]%
        {yuan2013time}
\bibfield{author}{\bibinfo{person}{Quan Yuan}, \bibinfo{person}{Gao Cong}, \bibinfo{person}{Zongyang Ma}, \bibinfo{person}{Aixin Sun}, {and} \bibinfo{person}{Nadia~Magnenat Thalmann}.} \bibinfo{year}{2013}\natexlab{}.
\newblock \showarticletitle{Time-aware point-of-interest recommendation}. In \bibinfo{booktitle}{\emph{Proceedings of the 36th international ACM SIGIR conference on Research and development in information retrieval}}. \bibinfo{pages}{363--372}.
\newblock


\bibitem[Zeng et~al\mbox{.}(2025)]%
        {zeng2025global}
\bibfield{author}{\bibinfo{person}{Jun Zeng}, \bibinfo{person}{Hongjin Tao}, \bibinfo{person}{Haoran Tang}, \bibinfo{person}{Junhao Wen}, {and} \bibinfo{person}{Min Gao}.} \bibinfo{year}{2025}\natexlab{}.
\newblock \showarticletitle{Global and local hypergraph learning method with semantic enhancement for POI recommendation}.
\newblock \bibinfo{journal}{\emph{Information Processing \& Management}} \bibinfo{volume}{62}, \bibinfo{number}{1} (\bibinfo{year}{2025}), \bibinfo{pages}{103868}.
\newblock


\bibitem[Zeng et~al\mbox{.}(2023)]%
        {zeng2023lgsa}
\bibfield{author}{\bibinfo{person}{Jun Zeng}, \bibinfo{person}{Yizhu Zhao}, \bibinfo{person}{Ziwei Wang}, \bibinfo{person}{Hongjin Tao}, \bibinfo{person}{Min Gao}, {and} \bibinfo{person}{Junhao Wen}.} \bibinfo{year}{2023}\natexlab{}.
\newblock \showarticletitle{LGSA: A next POI prediction method by using local and global interest with spatiotemporal awareness}.
\newblock \bibinfo{journal}{\emph{Expert Systems with Applications}}  \bibinfo{volume}{227} (\bibinfo{year}{2023}), \bibinfo{pages}{120291}.
\newblock


\bibitem[Zhao et~al\mbox{.}(2020)]%
        {zhao2020go}
\bibfield{author}{\bibinfo{person}{Pengpeng Zhao}, \bibinfo{person}{Anjing Luo}, \bibinfo{person}{Yanchi Liu}, \bibinfo{person}{Jiajie Xu}, \bibinfo{person}{Zhixu Li}, \bibinfo{person}{Fuzhen Zhuang}, \bibinfo{person}{Victor~S Sheng}, {and} \bibinfo{person}{Xiaofang Zhou}.} \bibinfo{year}{2020}\natexlab{}.
\newblock \showarticletitle{Where to go next: A spatio-temporal gated network for next poi recommendation}.
\newblock \bibinfo{journal}{\emph{IEEE Transactions on Knowledge and Data Engineering}} \bibinfo{volume}{34}, \bibinfo{number}{5} (\bibinfo{year}{2020}), \bibinfo{pages}{2512--2524}.
\newblock


\bibitem[Zhao et~al\mbox{.}(2024)]%
        {zhao2024let}
\bibfield{author}{\bibinfo{person}{Yuyue Zhao}, \bibinfo{person}{Jiancan Wu}, \bibinfo{person}{Xiang Wang}, \bibinfo{person}{Wei Tang}, \bibinfo{person}{Dingxian Wang}, {and} \bibinfo{person}{Maarten de Rijke}.} \bibinfo{year}{2024}\natexlab{}.
\newblock \showarticletitle{Let me do it for you: Towards llm empowered recommendation via tool learning}. In \bibinfo{booktitle}{\emph{Proceedings of the 47th International ACM SIGIR Conference on Research and Development in Information Retrieval}}. \bibinfo{pages}{1796--1806}.
\newblock


\bibitem[Zhong et~al\mbox{.}(2024)]%
        {zhong2024scfl}
\bibfield{author}{\bibinfo{person}{Lin Zhong}, \bibinfo{person}{Jun Zeng}, \bibinfo{person}{Ziwei Wang}, \bibinfo{person}{Wei Zhou}, {and} \bibinfo{person}{Junhao Wen}.} \bibinfo{year}{2024}\natexlab{}.
\newblock \showarticletitle{SCFL: Spatio-temporal consistency federated learning for next POI recommendation}.
\newblock \bibinfo{journal}{\emph{Information Processing \& Management}} \bibinfo{volume}{61}, \bibinfo{number}{6} (\bibinfo{year}{2024}), \bibinfo{pages}{103852}.
\newblock


\bibitem[Zhuang et~al\mbox{.}(2023)]%
        {zhuang2023toolqa}
\bibfield{author}{\bibinfo{person}{Yuchen Zhuang}, \bibinfo{person}{Yue Yu}, \bibinfo{person}{Kuan Wang}, \bibinfo{person}{Haotian Sun}, {and} \bibinfo{person}{Chao Zhang}.} \bibinfo{year}{2023}\natexlab{}.
\newblock \showarticletitle{Toolqa: A dataset for llm question answering with external tools}.
\newblock \bibinfo{journal}{\emph{Advances in Neural Information Processing Systems}}  \bibinfo{volume}{36} (\bibinfo{year}{2023}), \bibinfo{pages}{50117--50143}.
\newblock


\end{thebibliography}


\end{document}